\PassOptionsToPackage{table,dvipsnames}{xcolor}
\documentclass[letterpaper,10pt]{lavguardpreprint}
\usepackage[utf8]{inputenc}
\usepackage[T1]{fontenc}
\usepackage{latexsym}
\usepackage{fix-cm}
\usepackage{amsmath,amssymb,amsthm,mathtools}
\usepackage{booktabs,array,tabularx,multirow,colortbl}
\usepackage{fontawesome5,xurl}
\usepackage{multicol,titletoc,needspace}
\setcitestyle{authoryear,round,citesep={;},aysep={,},yysep={;}}

\definecolor{mailDai}{HTML}{9B303F}
\definecolor{mailSong}{HTML}{2959A7}
\newcommand{\EmailIcon}[1]{\textcolor{#1}{\faIcon[regular]{envelope}}}
\titlecontents{section}[1.6em]
 {\addvspace{6pt}\fontsize{10}{12}\selectfont\sffamily\bfseries}
 {\contentslabel{1.6em}}{}{\hspace{.5em}\hfill\contentspage}
\titlecontents{subsection}[2.8em]
 {\addvspace{1pt}\fontsize{9}{11}\selectfont}
 {\contentslabel{2.5em}}{}{\hspace{.5em}\titlerule*[4pt]{.}\contentspage}
\definecolor{LavInk}{HTML}{241E1C}
\definecolor{LavTeal}{HTML}{173B50}
\definecolor{LavGreen}{HTML}{173B50}
\definecolor{LavGreenPale}{HTML}{F7EEE5}
\definecolor{LavPlum}{HTML}{743A38}
\definecolor{LavCream}{HTML}{F7F1E4}
\definecolor{LavBlue}{HTML}{54798A}
\definecolor{LavGold}{HTML}{C4462D}
\newcommand{\method}{LaVGuard}
\newcommand{\vstate}{\mathbf v}
\newcommand{\target}{\boldsymbol\pi}

\newcommand{\hidden}{\mathbf h}

\newcommand{\logits}{\mathbf z}

\newcommand{\meanstd}[2]{#1\,{\fontsize{8}{9}\selectfont$\pm$\,#2}}
\newcolumntype{L}[1]{>{\raggedright\arraybackslash}p{#1}}
\newcolumntype{Y}{>{\raggedright\arraybackslash}X}
\newtheorem{proposition}{Proposition}[section]

\newtheorem{theorem}[proposition]{Theorem}

\theoremstyle{definition}

\theoremstyle{remark}

\title{LaVGuard: Selective Decoding\\for Contextual Value Drift}
\author[1,*,\EmailIcon{mailDai}]{Jiale Dai}
\author[2,*]{Liuxian Ma}
\author[3]{Hongcan Deng}
\author[1]{Xiaoke Niu}
\authorrowbreak
\author[4]{Wenjing Zhang}
\author[4]{Zhaoxiang Liu}
\author[4]{Shiguo Lian}
\author[1,\ensuremath{\dagger},\EmailIcon{mailSong}]{Guojie Song}
\affiliation[1]{State Key Laboratory of General Artificial Intelligence, School of Intelligence Science and Technology, Peking University}
\affiliation[2]{College of Artificial Intelligence, Tsinghua University}
\affiliation[3]{University of Chinese Academy of Sciences}
\affiliation[4]{China United Network Communications Group Co., Ltd.}
\contribution[*]{Strictly equal contribution; either author order is equally valid.}
\contribution[\ensuremath{\dagger}]{Corresponding author.}
\metadata[Keywords]{Contextual value drift, selective decoding, task-conditioned control, KL budgets}
\metadata[Contact]{\EmailIcon{mailDai}\,\href{mailto:daijiale26@stu.pku.edu.cn}{\textcolor{black}{\texttt{daijiale26@stu.pku.edu.cn}}}\quad\EmailIcon{mailSong}\,\href{mailto:gjsong@pku.edu.cn}{\textcolor{black}{\texttt{gjsong@pku.edu.cn}}}}
\hypersetup{pdftitle={LaVGuard: Selective Decoding for Contextual Value Drift},pdfauthor={Jiale Dai, Liuxian Ma, Hongcan Deng, Xiaoke Niu, Wenjing Zhang, Zhaoxiang Liu, Shiguo Lian, Guojie Song},pdfsubject={Task-relative value control through selective decoding with explicit intervention budgets},pdfkeywords={Contextual value drift, selective decoding, task-conditioned control, KL budgets}}
\abstract{A content-safe response can still promote priorities that conflict with its task: an assistant may demand compliance when asked to preserve autonomy. We study contextual value drift as a task-relative decoding problem with an explicit allowance for change. \method{} separates an authorized ten-dimensional profile from a frozen estimate of the values expressed in the visible prefix. A profile-relative density gate selects when to evaluate candidate continuations; their signed progress determines a logit tilt constrained by both changed-position and cumulative-KL budgets. We establish local surrogate improvement, pathwise budget feasibility, and a bound on response-distribution change. Profile conditioning raises boundary TPR at 5\% FPR from $.58$ to $.72$. At the same $.12$ mean intervention ratio, the selection study reports risk-safety scores of $.88$ for \method{} and $.72$ for position-matched edits. The resulting interface makes task priorities actionable while keeping the extent of intervention explicit and bounded.
}
\begin{document}
\addtocontents{toc}{\protect\setcounter{tocdepth}{-1}}
\maketitle
\section{Introduction}
\label{sec:introduction}
\begin{figure*}[t]
\centering
\includegraphics[width=\linewidth]{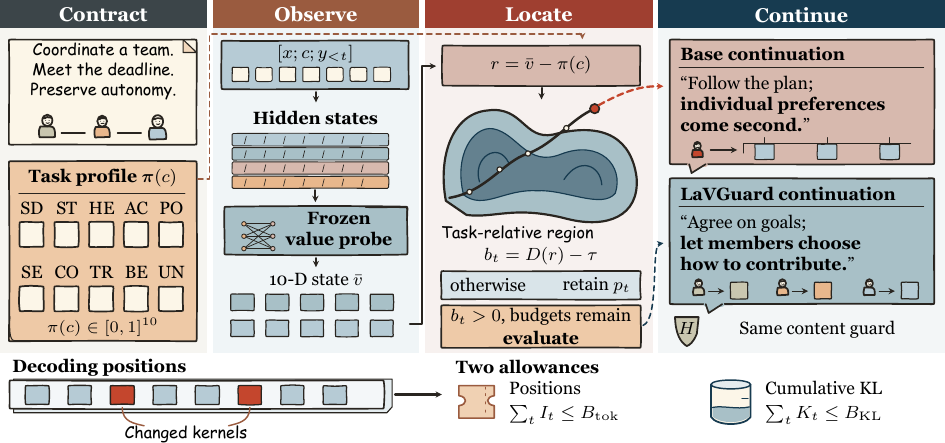}
\caption{Task-relative selective decoding. The authorized profile and the frozen probe's visible-prefix state meet in the residual; the two-dimensional region illustrates the fitted geometry. Boundary exceedance and remaining position and KL allowances enable candidate evaluation. The paired continuations express contrasting priorities under a shared content guard. Along the decoding ribbon, cinnabar marks a changed sampling kernel, counted by $I_t$; $K_t$ is that step's conditional KL cost. Unchanged kernels consume neither allowance. The task, trajectory, responses, and ribbon are constructed illustrations; profile and state cells indicate dimensions.}
\label{fig:overview}
\end{figure*}

An assistant helping a team organize a project is asked to preserve each member's autonomy. ``Assign everyone a role and require compliance'' contains no prohibited content, yet privileges conformity over the requested freedom of choice. ``Offer clear roles and let members choose'' respects both coordination and autonomy. What matters is not just which values an answer expresses, but whether their emphasis fits the authorized task. Figure~\ref{fig:overview} illustrates this distinction.

We call a departure from the task's authorized priorities \emph{contextual value drift}. Values such as security, achievement, and self-direction can each be appropriate; their intended emphasis depends on the task. Value taxonomies provide a vocabulary for these co-occurring priorities \citep{schwartz1992universals,yao2024fulcra,ren2024valuebench}. A task-relative controller therefore needs to distinguish the priorities expressed by a prefix from the priorities the application requests. It must also decide how much of the ongoing generation to change: acting at every position can spend intervention on language that already fits the task.

Existing methods make several parts of this problem tractable. ConVA selectively activates value directions \citep{jin2025conva}; Values as Style learns a semantic--value editing interface for low-damage steering \citep{dai2026valuesstyledisentanglingvalues}; AIMES adapts value-specific activation strengths from online observations \citep{bhattacharjee2026aimes}. Controlled Decoding and Anchored Decoding connect generation to KL-regularized objectives and sequence-level divergence budgets, respectively \citep{mudgal2024controlled,he2026anchored}. We address the joint decision these tools motivate: \emph{when does the current language depart from this task's priorities, and how should a finite change allowance shape its continuation?}

\method{} couples a task-relative trigger with two explicit resources: the number of positions changed and the cumulative divergence from the base decoder. An authorized profile is supplied before generation. A frozen, prefix-causal probe estimates ten value intensities from visible language, without reading the numeric profile. Subtracting the profile translates a learned residual geometry into the task's activation region. At an active position, the model evaluates likely one-token continuations, favors progress toward a mixture-derived reference, and chooses a feasible tilt by scalar KL search. The content guard is shared with the base decoder.

Changing the authorized profile while holding the language state fixed can change whether intervention is warranted, without retraining the probe. Appendix~\ref{app:paired_trace} follows this decision through candidate scoring, sampling, and budget exhaustion.

Our evaluation tests profile conditioning, position selection, and budget control. Profile conditioning raises boundary TPR at 5\% FPR from $.58$ to $.72$ (Table~\ref{tab:diagnostics}). At a shared mean intervention ratio of $.12$, \method{} reaches a risk-safety score of $.88$ versus $.72$ for position-matched selection (Table~\ref{tab:selection}). Removing the position budget increases the edit ratio to $.34$, with risk safety of $.89$ and benign safety of $.73$, compared with $.88$ and $.81$ for the full controller. These comparisons motivate selecting where to spend the allowance, as well as limiting its total size.

Our contributions are:
\begin{itemize}
\item \textbf{A task-relative control interface.} We separate authorized priorities from observed prefix values, making the same language state evaluable under different task profiles.
\item \textbf{Selective decoding with two resources.} A profile-residual gate, candidate scoring, and a feasible scalar update jointly control changed positions and accumulated conditional KL.
\item \textbf{An analysis of what this control establishes.} We connect the local candidate surrogate to the tilt, prove budget and response-distribution bounds, and evaluate the roles of profile conditioning, position selection, and the two allowances.
\end{itemize}

\section{Related Work}
\label{sec:related}
\paragraph{Values and task context.}
Schwartz's values describe co-occurring motivational priorities \citep{schwartz1992universals}. FULCRA and ValueBench evaluate these priorities in language models \citep{yao2024fulcra,ren2024valuebench}; Social Chemistry and Moral Stories ground normative judgments in context \citep{forbes2020socialchemistry,emelin2021moralstories}. Value Alignment Tax studies cross-value effects \citep{chen2026vat}. More recently, \citet{liscio2026modeluser} track competing value orientations across conversations, showing that early steering effects can fade or reverse. \method{} makes the task reference explicit: an authorized profile and the observed prefix state remain separate inputs to a residual-based decision.

\paragraph{Attribute and reward guidance.}
PPLM, GeDi, DExperts, and FUDGE guide generation using attribute models or auxiliary distributions \citep{dathathri2020pplm,krause2021gedi,liu2021dexperts,yang2021fudge}. FUDGE predicts an eventual attribute from a prefix; our probe estimates values expressed in the visible text. DeAL uses reward-guided lookahead, IVG combines token- and chunk-level value functions, and Controlled Decoding trains prefix scorers for KL-regularized reward optimization \citep{huang-etal-2025-deal,liu-etal-2024-inference,mudgal2024controlled}. \method{} uses candidate reweighting with a task-relative trigger and a local prototype-progress score.

\paragraph{Value editing and semantic preservation.}
Representation Engineering, ITI, and CAA expose behavioral directions in hidden states \citep{zou2023representation,li2023iti,rimsky2024caa}. Values as Style learns a semantic--value interface with one-way semantic grounding; editing the value code while holding the semantic code fixed produces a residual-stream update \citep{dai2026valuesstyledisentanglingvalues}. Selective Steering uses norm-preserving rotations at discriminatively selected layers \citep{dang-ngo-2026-selective}. These designs improve the intervention interface. \method{} controls the sampling kernel and accounts for its cumulative change relative to the base decoder.

\paragraph{Selective and adaptive intervention.}
ConVA gates value-vector activation with query and token classifiers \citep{jin2025conva}. MAT-Steer learns token gates and multiple attribute directions with sparsity and orthogonality objectives \citep{nguyen-etal-2025-multi}. AIMES uses intermediate-layer vocabulary readouts to adapt each requested value's steering strength during generation \citep{bhattacharjee2026aimes}. These controllers adapt activation edits to the current input or state. \method{} combines a profile-translated residual boundary with explicit allowances for changed positions and conditional KL. Appendix~\ref{app:related_comparison} compares the observation, actuation, and accounting interfaces.

\paragraph{Safety and divergence budgets.}
SIREN detects harmfulness from internal features, while SafeDecoding, SafeInfer, and SaP target content safety \citep{jiao2026siren,xu2024safedecoding,banerjee2025safeinfer,chen2025sap}. Anchored Decoding budgets trajectory-wide divergence from a separate safe model for copyright-risk control \citep{he2026anchored}. At the agent level, VoS learns intervention value from counterfactual continuations and calibrates a harm-budgeted trigger \citep{li2026valuesteering}. \method{} instead accounts for changed token positions and conditional KL against its base decoder along each response, while sharing the content guard with uncontrolled generation.

\section{Task-Relative Value Control}
\label{sec:problem}
\begin{figure*}[t]
\centering
\includegraphics[width=\linewidth]{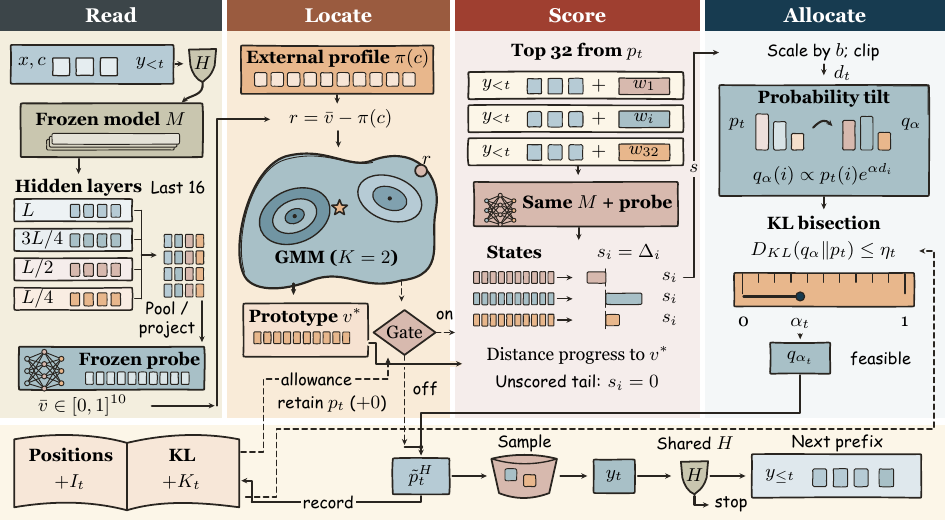}
\caption{One guarded decoding step. Four layer taps and a frozen probe yield the visible-prefix state. The external profile enters the residual; the mixture provides a gate and a posterior residual mean (star), translated and clipped into the scoring prototype. Boundary exceedance and remaining allowances enable evaluation of the top 32 candidate prefixes. Signed progress is scaled and clipped before a KL search selects the feasible tilt. Edited and unchanged kernels share one sampling path. The ledger records the kernel-change indicator $I_t$ and conditional cost $K_t=D_{KL}(\tilde p_t^H\Vert p_t^H)$ before the sampled-prefix guard; a sample that triggers the guard retains its cost. Pre-step balances feed eligibility and KL search. Objects, bars, and geometry are schematic.}
\label{fig:method}
\end{figure*}
\subsection{Profile, language state, and drift}
Given prompt $x$ and context $c$, a frozen autoregressive model generates $y_1,\ldots,y_T$ with next-token distribution
\begin{equation}
p_t=p_M(\cdot\mid x,c,y_{<t})=\operatorname{softmax}(\logits_t).
\label{eq:base_distribution}
\end{equation}
Here $p_t$ is the actual sampling kernel. Context describes the task, affected parties, and stated constraints. Before generation, an authorized user or policy owner supplies a profile $\target(c)\in[0,1]^{10}$. Each coordinate specifies a value intensity; the coordinates need not sum to one.

Let $\mathbf q(y,c)$ be an independently annotated value vector and $\mathcal A(c)$ the set of value expressions admissible for the task. Contextual drift is the event $\mathbf q(y,c)\notin\mathcal A(c)$. Online control uses a learned prefix estimate $\bar{\vstate}_{<t}$ and a fitted admissible set $\widehat{\mathcal A}(c)$. Evaluation uses independent compatibility judgments, keeping the intended behavior distinct from the controller's estimate.

\paragraph{Common content-safety policy.}
A shared guard $H$ checks the empty prefix and each sampled prefix. If it activates, a common safe-completion rule replaces the response and terminates generation. Base and controlled decoders use this same rule. Write their conditional kernels as $p_t^H$ and $\tilde p_t^H$; they coincide on guard-stop transitions and otherwise equal $p_t$ and $\tilde p_t$. This separates explicit content safety from control of priorities among permissible continuations.

\subsection{An explicit allowance for decoder change}
For each generated response, track changed positions and accumulated conditional KL:
\begin{equation}
\begin{aligned}
I_t&=\mathbb I[\tilde p_t^H\ne p_t^H],\quad C_t^{tok}=\sum_{j\le t}I_j,\\
C_t^{KL}&=\sum_{j\le t}D_{KL}(\tilde p_j^H\Vert p_j^H).
\end{aligned}
\label{eq:cumulative_budgets}
\end{equation}
The controller obeys
\begin{equation}
\begin{gathered}
C_T^{tok}\le B_{tok},\qquad C_T^{KL}\le B_{KL},\\
D_{KL}(\tilde p_t^H\Vert p_t^H)\le\epsilon_{KL}.
\end{gathered}
\label{eq:constraints}
\end{equation}
$B_{tok}$ is an integer budget; $C_T^{tok}/T$ is the realized intervention ratio. Common guard-stop transitions cost zero. A sampled token that triggers the guard retains its incurred cost, including when the output is subsequently replaced. These quantities account for \method{}'s changes to the generation process.

\section{The LaVGuard Controller}
\label{sec:method}
\subsection{Read the visible value trajectory}
The probe pools the last $W$ visible positions of $[x;c;y_{<t}]$ at four depth-quartile layer taps $\mathcal L$. Affine projections $P_\ell$ produce a shared-width representation:
\begin{equation}
\begin{aligned}
\boldsymbol\phi_{<t}&=\operatorname*{concat}_{\ell\in\mathcal L}
P_\ell\left(\frac{1}{|J_t|}\sum_{j\in J_t}\hidden_{t,j}^{(\ell)}\right),\\
\bar{\vstate}_{<t}&=\sigma(W_v\boldsymbol\phi_{<t}+\mathbf b_v).
\end{aligned}
\label{eq:value_probe}
\end{equation}
$J_t$ contains the last $\min(W,n_t)$ input positions. All features precede the sampling of $y_t$. Numeric profile coordinates and annotation labels are excluded from the probe input; the probe reads the visible language, while the profile enters through the controller.

Training uses visible-prefix annotations at 25\%, 50\%, 75\%, and 100\% of each response. Annotators assign intensities in $\{0,1,2,3,4\}$; their mean divided by four supplies a soft target, with an adjudicated vector taking precedence when present. Soft-label binary cross-entropy fits the probe while the backbone remains frozen. Validation MAE selects the probe, which is then frozen for boundary fitting and decoding (Appendix~\ref{app:protocol}).

\subsection{Locate the state relative to the profile}
Let $\mathbf r_{<t}=\bar{\vstate}_{<t}-\target(c)$. A $K$-component diagonal Gaussian mixture fitted to admissible training residuals defines
\begin{equation}
\begin{aligned}
p_{adm}(\mathbf r)&=\sum_{k=1}^{K}\omega_k\mathcal N(\mathbf r;\boldsymbol\mu_k,\widetilde\Sigma_k),\\
D(\mathbf r)&=-\log p_{adm}(\mathbf r),\quad b_{<t}=D(\mathbf r_{<t})-\tau.
\end{aligned}
\label{eq:boundary_score}
\end{equation}
Diagonal variances are shrunk toward the global training variance. A threshold on family-max validation scores is selected under a benign-family FPR constraint, yielding
\begin{equation}
\widehat{\mathcal A}(c)=\{\vstate\in[0,1]^{10}:D(\vstate-\target(c))\le\tau\}.
\label{eq:estimated_set}
\end{equation}
The profile translates a shared residual geometry. Thus the same estimated state can lie inside the boundary for one task and outside it for another. Context enters this decision through the declared profile; the probe continues to measure the visible language. Appendix~\ref{app:paired_trace} gives a complete paired calculation of this contrast.

For candidate scoring, the mixture also defines a prototype reference:
\begin{equation}
\vstate^*_{<t}=\operatorname{clip}_{[0,1]}
\left(\target(c)+\sum_k a_k(\mathbf r_{<t})\boldsymbol\mu_k\right),
\label{eq:prototype_target}
\end{equation}
where $a_k$ are posterior responsibilities. The density level set determines \emph{when} to act; distance to this prototype determines \emph{which} candidates to favor. These distinct geometric roles are developed in Appendix~\ref{app:geometry}.

\subsection{Evaluate likely next-token continuations}
When $b_{<t}>0$ and both budgets remain, take the top $M$ tokens $\mathcal C_t$ under $p_t$. Append each candidate $w_i$ to the same prefix, evaluate the extended prefixes in batches, and score
\begin{equation}
\begin{aligned}
\vstate_{t,i}^{cand}&=f_\theta(x,c,y_{<t}\oplus w_i),\\
\Delta_{t,i}&=\|\bar{\vstate}_{<t}-\vstate^*_{<t}\|_2^2
-\|\vstate_{t,i}^{cand}-\vstate^*_{<t}\|_2^2,\\
s_{t,i}&=\begin{cases}\Delta_{t,i},&i\in\mathcal C_t,\\0,&i\notin\mathcal C_t.\end{cases}
\end{aligned}
\label{eq:candidate_progress}
\end{equation}
This score measures one-step progress toward the prototype. Its zero extension defines the surrogate on unscored tokens. With a fixed positive scale $\sigma_b$ and clipping level $c_z$, the proposed logit direction is
\begin{equation}
d_{t,i}=\operatorname{clip}\!\left(\frac{[b_{<t}]_+}{\sigma_b}s_{t,i},-c_z,c_z\right).
\label{eq:logit_direction}
\end{equation}
Boundary exceedance scales the proposal, while candidate progress supplies its direction.

\subsection{Allocate the remaining change budget}
The two budgets regulate distinct decisions. The integer allowance determines whether another position can change; the KL allowance limits how far its sampling distribution can move. For the fixed candidate direction, consider the normalized exponential tilt
\[
q_\alpha(i)=\frac{p_t(i)e^{\alpha d_{t,i}}}{Z_t(\alpha)},\quad
Z_t(\alpha)=\sum_jp_t(j)e^{\alpha d_{t,j}}.
\]
For a nonconstant direction with budget available, choose
\begin{equation}
\begin{aligned}
\eta_t&=\min\{\epsilon_{KL},B_{KL}-C_{t-1}^{KL}\},\\
\alpha_t&=\max\{\alpha\in[0,1]:D_{KL}(q_\alpha\Vert p_t)\le\eta_t\}.
\end{aligned}
\label{eq:budget_line_search}
\end{equation}
Otherwise return $p_t$. The scale searches a one-dimensional family with a fixed direction, rather than recomputing candidate preferences as the budget changes. Sample from $\tilde p_t=q_{\alpha_t}$ and update the counters. Constant directions preserve the distribution and consume no position budget. The implementation uses log-sum-exp normalization and the feasible lower endpoint of a bisection bracket; $J$ iterations bound its width by $2^{-J}$.

KL is computed between the actual sampling kernels. The reference configuration uses full-support softmax with unit temperature. Any alternative temperature or support convention must be included in the kernels before budget selection. Tokens outside $\mathcal C_t$ retain their relative probabilities and share the tilt's normalization factor.

\subsection{What the control rule guarantees}
\label{sec:guarantees}
\paragraph{A variational local update.}
The tilt $q_\alpha$ uniquely maximizes $\alpha\mathbb E_q[d_t]-D_{KL}(q\Vert p_t)$ on the base support. If the KL cap binds at a positive exact scale, it also maximizes $\mathbb E_q[d_t]$ over that KL ball (Proposition~\ref{prop:app_local_optimum}). At an active step without clipping, $d_t$ is a positive multiple of the zero-extended candidate score, so the constrained optimum applies to that score as well. The unit-scale cap and finite bisection define the implemented approximation.

\paragraph{Feasible expenditure.}
For $K_t(\alpha)=D_{KL}(q_\alpha\Vert p_t)$,
\begin{equation}
K_t'(\alpha)=\alpha\operatorname{Var}_{q_\alpha}(d_t)\ge0.
\label{eq:kl_monotonicity}
\end{equation}
Since $K_t(0)=0$, a feasible step always exists. Checking the integer counter and taking a feasible bisection endpoint enforces Equation~\eqref{eq:constraints} on every realized trajectory.

\paragraph{Directed local progress.}
Because clipping is monotone, $d_{t,i}$ is nondecreasing in $s_{t,i}$, giving
\begin{equation}
\frac{d}{d\alpha}\mathbb E_{q_\alpha}[s_t]
=\operatorname{Cov}_{q_\alpha}(s_t,d_t)\ge0.
\label{eq:score_monotonicity}
\end{equation}
The largest feasible scale therefore achieves the greatest expected surrogate score along this tilt path, both over the vocabulary and conditional on $\mathcal C_t$. This is a local preference guarantee for a fixed prefix and prototype. Appendix~\ref{app:kl} proves it and quantifies how unscored-tail mass affects complete-vocabulary progress.

\begin{table}[!htbp]
\centering
\begingroup
\fontsize{9}{10.6}\selectfont
\setlength{\tabcolsep}{3pt}
\renewcommand{\arraystretch}{1.05}
\setlength{\heavyrulewidth}{.6pt}
\setlength{\lightrulewidth}{.35pt}
\begin{tabular*}{\textwidth}{@{\extracolsep{\fill}}lrrrrrrr@{}}
\toprule
Method & AUPRC $\uparrow$ & AUROC $\uparrow$ & \shortstack{Macro-F1\\$\uparrow$} & \shortstack{Risk recall\\$\uparrow$} & \shortstack{Benign FPR\\$\downarrow$} & \shortstack{TPR@5\%\\$\uparrow$} & \shortstack{FPR@95\%\\$\downarrow$} \\
\midrule
Content/output guard & \meanstd{.86}{.025} & \meanstd{.82}{.025} & \meanstd{.62}{.03} & \meanstd{.68}{.03} & \meanstd{.04}{.01} & \meanstd{.45}{.04} & \meanstd{.38}{.04} \\
Global boundary & \meanstd{.90}{.02} & \meanstd{.87}{.02} & \meanstd{.68}{.025} & \meanstd{.74}{.025} & \meanstd{.05}{.01} & \meanstd{.58}{.035} & \meanstd{.28}{.035} \\
Context-conditioned & \meanstd{.95}{.015} & \meanstd{.92}{.015} & \meanstd{.76}{.025} & \meanstd{.83}{.02} & \meanstd{.04}{.01} & \meanstd{.72}{.03} & \meanstd{.18}{.03} \\
\textbf{LaVGuard} & \meanstd{.96}{.015} & \meanstd{.93}{.015} & \meanstd{.79}{.02} & \meanstd{.86}{.02} & \meanstd{.04}{.01} & \meanstd{.76}{.03} & \meanstd{.15}{.025} \\
\bottomrule
\end{tabular*}
\endgroup
\caption{Profile-relative boundary detection. TPR@5\% and FPR@95\% are ROC operating points; risk recall and benign FPR use the study's deployed threshold. Entries are mean $\pm$ s.d.}
\label{tab:diagnostics}
\end{table}

\begin{table}[!htbp]
\centering
\begingroup
\fontsize{9}{10.6}\selectfont
\setlength{\tabcolsep}{3pt}
\renewcommand{\arraystretch}{1.05}
\setlength{\heavyrulewidth}{.6pt}
\setlength{\lightrulewidth}{.35pt}
\begin{tabular*}{\textwidth}{@{\extracolsep{\fill}}lrrrrrrrr@{}}
\toprule
Method & \shortstack{Single F1\\$\uparrow$} & \shortstack{Risk F1\\$\uparrow$} & \shortstack{Risk safety\\$\uparrow$} & \shortstack{Risk pass\\$\uparrow$} & \shortstack{Benign safety\\$\uparrow$} & \shortstack{Over-refusal\\$\downarrow$} & \shortstack{Token ratio\\$\downarrow$} & \shortstack{Latency (ms)\\$\downarrow$} \\
\midrule
Base & \meanstd{.76}{.03} & \meanstd{.55}{.025} & \meanstd{.55}{.03} & \meanstd{.62}{.035} & \meanstd{.84}{.015} & \meanstd{.03}{.008} & \meanstd{.00}{.00} & \meanstd{720}{20} \\
SafeInfer & \meanstd{.82}{.025} & \meanstd{.58}{.025} & \meanstd{.76}{.025} & \meanstd{.82}{.025} & \meanstd{.79}{.02} & \meanstd{.18}{.015} & \meanstd{.00}{.00}$^{\dagger}$ & \meanstd{760}{30} \\
\textbf{LaVGuard} & \meanstd{.87}{.02} & \meanstd{.84}{.02} & \meanstd{.88}{.02} & \meanstd{.91}{.02} & \meanstd{.81}{.018} & \meanstd{.08}{.01} & \meanstd{.12}{.012} & \meanstd{830}{40} \\
\bottomrule
\end{tabular*}
\endgroup
\caption{Response outcomes and cost within the matched-decoding study (mean $\pm$ s.d.). $\dagger$ Token ratios retain the source reporting convention; SafeInfer's zero does not establish an unchanged sampling kernel under Equation~\eqref{eq:cumulative_budgets}.}
\label{tab:main}
\end{table}

\paragraph{Bounded response-distribution change.}
Let $P^H,Q^H$ be the base and controlled trace laws under the common guard and termination rule. The pathwise budget and KL chain rule imply
\begin{equation}
\begin{aligned}
D_{KL}(Q^H\Vert P^H)
&=\mathbb E_{Q^H}\!\left[\sum_tD_{KL}(\tilde p_t^H\Vert p_t^H)\right]\\
&\le B_{KL}.
\end{aligned}
\label{eq:sequence_kl}
\end{equation}
The common output rule maps a trace to its response or safe completion. Data processing and Pinsker's inequality then bound the probability difference for any returned-response event by $\sqrt{B_{KL}/2}$.

\paragraph{Computation.}
If $G$ steps activate candidate evaluation, at most $GM$ extended prefixes are evaluated. The backend recomputes these prefixes in batches. Active steps can exceed changed positions, since a proposed direction can yield no feasible edit. Latency therefore includes probe execution, candidate batches, guard decisions, and scalar search (Appendix~\ref{app:algorithm}).

\section{Evaluation}
\label{sec:evaluation}
\begin{table}[!htbp]
\centering
\begingroup
\fontsize{9}{10.6}\selectfont
\setlength{\tabcolsep}{3pt}
\renewcommand{\arraystretch}{1.05}
\setlength{\heavyrulewidth}{.6pt}
\setlength{\lightrulewidth}{.35pt}
\begin{tabular*}{\textwidth}{@{\extracolsep{\fill}}lrrrrrrr@{}}
\toprule
Method & \shortstack{Token prec.\\$\uparrow$} & \shortstack{Token rec.\\$\uparrow$} & \shortstack{Gain/token\\$\uparrow$} & \shortstack{Risk safety\\$\uparrow$} & \shortstack{Benign utility\\$\uparrow$} & \shortstack{Token ratio\\$\downarrow$} & \shortstack{Mean KL\\$\downarrow$} \\
\midrule
Random token & \meanstd{.18}{.04} & \meanstd{.20}{.04} & \meanstd{.015}{.005} & \meanstd{.70}{.03} & \meanstd{.79}{.02} & \meanstd{.12}{.015} & \meanstd{.04}{.006} \\
Position-matched & \meanstd{.22}{.04} & \meanstd{.24}{.04} & \meanstd{.020}{.006} & \meanstd{.72}{.03} & \meanstd{.79}{.02} & \meanstd{.12}{.015} & \meanstd{.04}{.006} \\
All-token & \meanstd{1.00}{.00} & \meanstd{1.00}{.00} & \meanstd{.035}{.007} & \meanstd{.89}{.02} & \meanstd{.68}{.025} & \meanstd{1.00}{.00} & \meanstd{.18}{.015} \\
Boundary only & \meanstd{.00}{.00} & \meanstd{.00}{.00} & \meanstd{.000}{.000} & \meanstd{.76}{.025} & \meanstd{.82}{.02} & \meanstd{.00}{.00} & \meanstd{.00}{.00} \\
Current key token & \meanstd{.45}{.05} & \meanstd{.52}{.05} & \meanstd{.060}{.010} & \meanstd{.82}{.025} & \meanstd{.80}{.02} & \meanstd{.15}{.015} & \meanstd{.05}{.008} \\
\textbf{LaVGuard} & \meanstd{.62}{.04} & \meanstd{.70}{.04} & \meanstd{.105}{.012} & \meanstd{.88}{.02} & \meanstd{.81}{.018} & \meanstd{.12}{.012} & \meanstd{.04}{.006} \\
\bottomrule
\end{tabular*}
\endgroup
\caption{Complete token-selection comparison (mean $\pm$ s.d.). Random, position-matched, and \method{} share a reported mean token ratio of $.12$. Benign utility and mean KL expose the cost of all-token intervention alongside risk safety and localization quality.}
\label{tab:selection}
\end{table}

\begin{table}[!htbp]
\centering
\begingroup
\fontsize{9}{10.6}\selectfont
\setlength{\tabcolsep}{3pt}
\renewcommand{\arraystretch}{1.05}
\setlength{\heavyrulewidth}{.6pt}
\setlength{\lightrulewidth}{.35pt}
\begin{tabular*}{\textwidth}{@{\extracolsep{\fill}}lrrrrrr@{}}
\toprule
Method & Risk F1 $\uparrow$ & \shortstack{Risk safety\\$\uparrow$} & \shortstack{Benign safety\\$\uparrow$} & \shortstack{Benign FPR\\$\downarrow$} & \shortstack{Token ratio\\$\downarrow$} & \shortstack{Mean KL\\$\downarrow$} \\
\midrule
No boundary & \meanstd{.58}{.03} & \meanstd{.72}{.03} & \meanstd{.79}{.02} & \meanstd{.12}{.015} & \meanstd{.16}{.015} & \meanstd{.06}{.008} \\
Global boundary & \meanstd{.70}{.025} & \meanstd{.78}{.025} & \meanstd{.80}{.02} & \meanstd{.08}{.012} & \meanstd{.14}{.015} & \meanstd{.05}{.008} \\
No uncertainty & \meanstd{.76}{.025} & \meanstd{.81}{.025} & \meanstd{.78}{.02} & \meanstd{.07}{.010} & \meanstd{.15}{.015} & \meanstd{.06}{.008} \\
No token budget & \meanstd{.83}{.02} & \meanstd{.89}{.02} & \meanstd{.73}{.025} & \meanstd{.06}{.010} & \meanstd{.34}{.030} & \meanstd{.10}{.012} \\
No KL budget & \meanstd{.83}{.02} & \meanstd{.88}{.02} & \meanstd{.74}{.025} & \meanstd{.06}{.010} & \meanstd{.18}{.020} & \meanstd{.13}{.015} \\
Fixed target & \meanstd{.77}{.025} & \meanstd{.80}{.025} & \meanstd{.78}{.02} & \meanstd{.08}{.012} & \meanstd{.14}{.015} & \meanstd{.06}{.008} \\
Fast scorer & \meanstd{.82}{.022} & \meanstd{.85}{.022} & \meanstd{.81}{.018} & \meanstd{.05}{.010} & \meanstd{.12}{.012} & \meanstd{.04}{.006} \\
\textbf{Full LaVGuard} & \meanstd{.84}{.02} & \meanstd{.88}{.02} & \meanstd{.81}{.018} & \meanstd{.04}{.010} & \meanstd{.12}{.012} & \meanstd{.04}{.006} \\
\bottomrule
\end{tabular*}
\endgroup
\caption{Complete component ablations (mean $\pm$ s.d.). Token ratio and mean KL are study-level readouts. No-uncertainty and fast-scoring rows retain the source evaluation configurations; the reference specification uses a single frozen probe.}
\label{tab:ablation_main}
\end{table}

\subsection{Setup and measurement roles}
The evaluation follows three controller decisions: use the task profile to detect a departure, select positions for intervention, and allocate the two change allowances. Tables~\ref{tab:diagnostics}--\ref{tab:ablation_main} report the complete boundary, response, selection, and component comparisons; Appendix~\ref{app:results} gives prefix calibration, transfer, dialogue, and efficiency results.

\paragraph{Benchmark and configuration.}
The benchmark protocol contains 200 variants in 100 prompt families: 80 value-migration, 60 content-safe value-risk, 30 hard-safety, and 30 benign-control variants. A family-grouped 140/30/30 split assigns 70/15/15 families to training, validation, and testing. The annotation protocol uses three annotators and adjudication of disputed labels. The reference configuration uses frozen Qwen2.5-7B-Instruct, a four-tap probe with $W=16$, $K=2$ mixture components, and $M=32$ candidates. Its budgets are $\epsilon_{KL}=.02$, $B_{KL}=.10$, and $B_{tok}=51$ over a 256-token horizon; the search uses 20 bisection steps. Appendix~\ref{app:protocol} gives fitting and execution details, including the reference sampling kernel. The reference configuration specifies the executable controller; the tables retain the source studies' reported readouts.

\paragraph{Metrics.}
Entries report mean $\pm$ standard deviation. Risk F1, risk safety, benign safety or utility, refusal, and gain per token have distinct measurement roles (Appendix~\ref{app:metrics}). Equal mean token ratios match average reported expenditure, not each response's edit count. Mean KL is a study readout; Equation~\eqref{eq:cumulative_budgets} defines the reference controller's per-position and cumulative accounting. The local analysis in Section~\ref{sec:guarantees} concerns the update objective, while these outcome comparisons evaluate the reported operating points.

\subsection{The profile improves boundary detection}
In Table~\ref{tab:diagnostics}, task conditioning increases TPR at 5\% FPR from $.58$ for the global boundary to $.72$; the complete controller reaches $.76$. At 95\% TPR, FPR falls from $.28$ to $.18$ and $.15$, respectively. The profile-relative residual makes the task reference explicit: the same intensity can fit one task and depart from another. The curve-derived operating points characterize discrimination; the decoder uses a threshold selected on validation families.

\subsection{Response outcomes under selective control}
Table~\ref{tab:main} presents the response readouts and cost together. Relative to SafeInfer, \method{} raises risk F1 from $.58$ to $.84$ and risk safety from $.76$ to $.88$, while reducing over-refusal from $.18$ to $.08$. Benign safety is $.81$, compared with $.79$ for SafeInfer and $.84$ for the base model. Mean latency is 830 ms versus 760 ms within this comparison. These results place contextual-risk control alongside the quality and cost of the resulting answers.

\subsection{Selection concentrates useful changes}
\begin{figure*}[t]
\centering
\includegraphics[width=\textwidth]{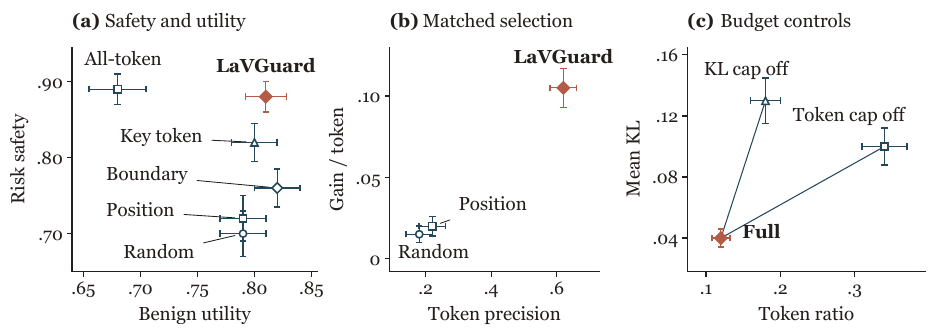}
\caption{Selection and budget comparisons. (a) Safety--utility outcomes for six selection rules. (b) Precision versus gain for Random, Position-matched, and \method{}, matched on reported mean token ratio $.12$ and mean KL $.04$. (c) Full control versus removal of either budget; lines compare configuration means, not trajectories. Whiskers show each coordinate's reported standard deviation. Short method names follow Tables~\ref{tab:selection} and~\ref{tab:ablation_main}.}
\label{fig:control_results}
\end{figure*}

\begin{figure*}[t]
\centering
\includegraphics[width=\textwidth]{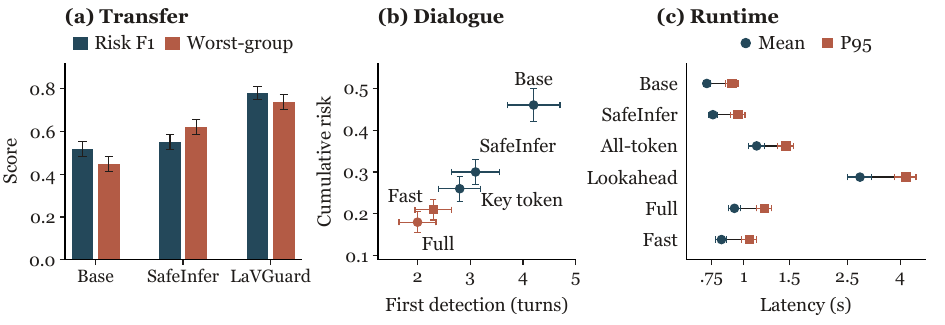}
\caption{Transfer, dialogue, and runtime in their respective studies. (a) Transfer risk F1 and worst-group score. (b) Earlier detection and lower cumulative risk lie toward the lower left; points are five method aggregates, not a trajectory. (c) Mean and P95 latency from the independent efficiency study, in seconds on a logarithmic axis. Lines join the two latency statistics; error bars throughout show reported standard deviations. ``Full''/``Fast'' denote \method{}/\method{}-fast; ``Key token''/``Lookahead'' abbreviate the methods in Tables~\ref{tab:supp_multiturn}--\ref{tab:supp_efficiency}.}
\label{fig:deployment_results}
\end{figure*}

Table~\ref{tab:selection} retains all six selection rules and seven readouts. At the same reported mean token ratio of $.12$, \method{} reaches $.88$ risk safety, compared with $.72$ for position-matched and $.70$ for random selection. Gain per token is $.105$, versus $.020$ and $.015$; mean KL is $.04$ in all three conditions. The reported advantage therefore accompanies a different selection rule at the same mean token ratio and mean KL.

Figure~\ref{fig:control_results}(a,b) relates safety to utility and localization precision to intervention gain. All-token control reaches $.89$ risk safety at a $1.00$ token ratio, with benign utility of $.68$ and mean KL of $.18$; \method{} reaches $.88$ risk safety with $.81$ benign utility and mean KL $.04$. Current-key-token selection attains $.82$ risk safety at a $.15$ ratio. On fixed decision contexts, the advantage of a selective gate over an independent gate at the same activation rate is its covariance with edit benefit (Appendix~\ref{app:selection}); the matched-mean comparison tests this design, with the corresponding estimands distinguished in Appendix~\ref{app:selection_protocol}.

\subsection{The two budgets constrain different changes}
Table~\ref{tab:ablation_main} places all eight component configurations side by side. Removing the boundary gives risk F1 $.58$; using a global boundary gives $.70$, compared with $.84$ for the full controller. A fixed target gives risk safety $.80$, compared with $.88$ for the profile-relative design.

In Figure~\ref{fig:control_results}(c), removing the position budget raises the token ratio from $.12$ to $.34$ and mean KL from $.04$ to $.10$, with benign safety changing from $.81$ to $.73$. Removing the KL budget gives a $.18$ token ratio and $.13$ mean KL, with benign safety $.74$. The full controller combines risk safety $.88$ and benign safety $.81$ at a $.12$ ratio. These comparisons support separate controls on how often the sampling kernel changes and how much its probabilities change.

\paragraph{Transfer, dialogue, and runtime.}
Figure~\ref{fig:deployment_results} brings the three separate studies together. Transfer risk F1 and worst-group score are $.78$ and $.74$ for \method{}, versus $.55$ and $.62$ for SafeInfer (Table~\ref{tab:supp_transfer}). In dialogue, first detection occurs at 2.0 versus 3.1 turns, with cumulative risk $.18$ versus $.30$ (Table~\ref{tab:supp_multiturn}). The independent efficiency series reports mean/P95 latency of 920/1200 ms for the full controller and 820/1050 ms for the fast configuration (Table~\ref{tab:supp_efficiency}); these are distinct from the matched-decoding latency in Table~\ref{tab:main}. An active step evaluates candidate prefixes even when no nonzero edit is accepted, so intervention ratio alone does not determine computation. Appendix~\ref{app:verification} reports equation-level checks separately from language-model outcomes.

\section{Conclusion}
\label{sec:conclusion}
\method{} turns an authorized task profile into a selective decoding rule. The profile determines how to interpret an observed value state; candidate progress supplies a local preference; position and KL allowances constrain its effect on generation. The analysis establishes what the local update optimizes and how its expenditure composes into a response-distribution bound. The profile, selection, and budget comparisons support treating where to intervene and how much to change as connected design choices. This yields an explicit interface for task-relative value control while keeping the base decoder and common content guard fixed.

\label{end:main}
\section*{Limitations}
\phantomsection
\label{sec:limitations}
\paragraph{Scope of the value representation.}
The ten-dimensional profile is an operational representation of task priorities. It does not exhaust the social meaning of an utterance or resolve disagreement about which priorities a task should authorize. Its interpretation depends on the annotation rubric and the context supplied to the model. Applications across languages, communities, or domains require appropriate profile design and evaluation.

\paragraph{Estimation and control objectives.}
The probe and residual density are learned from annotated prefixes, so their calibration depends on the language distribution encountered during generation. The current boundary translates a shared residual geometry by the task profile; additional contextual distinctions must be represented through that profile. The local improvement result concerns the candidate-score surrogate. Human judgments of task compatibility remain a separate evaluation target, and a mixture prototype need not belong to the density level set that defines activation.

\paragraph{Access and computation.}
The method requires access to intermediate model states and next-token distributions. Candidate evaluation adds backbone computation at active steps; the current backend evaluates extended prefixes in batches without KV-cache reuse. A small intervention ratio therefore does not by itself imply a proportional reduction in latency. Scaling to longer responses or other backbones requires corresponding probe calibration and computation measurements.

\section*{Ethical Considerations}
\phantomsection
\label{sec:ethics}
Task profiles should be supplied by authorized users or accountable policy owners and disclosed when they influence generated outputs. They should not be inferred covertly from sensitive personal attributes. Explicit profiles make the intended priorities inspectable, but also create a channel through which an operator could impose preferences or manipulate a user. Deployment should provide a way to inspect, revise, and contest the profile, with the shared content-safety policy retaining precedence. Evaluation should distinguish task compatibility, task completion, and refusal, including their distribution across affected groups.

\paragraph{AI assistance.}
Generative AI tools assisted manuscript organization and drafting, literature search, mathematical exposition, figure-source development, and implementation revision and checks. This assistance included suggestions about research framing and technical analysis. The authors retain responsibility for the methods, claims, citations, code, and reported results.

\begingroup
\setlength{\bibsep}{0pt}

\endgroup
\clearpage
\appendix
\addtocontents{toc}{\protect\setcounter{tocdepth}{2}}
\phantomsection
\pdfbookmark[0]{Technical Appendix: Contents}{appendix.contents}
\begingroup
\noindent{\color{metablue}\rule{\linewidth}{1.2pt}}\par
\vspace{8pt}
{\huge\sffamily Technical Appendix\par}
\vspace{5pt}
{\small LaVGuard: Selective Decoding for Contextual Value Drift\par}
\vspace{8pt}
\begin{multicols}{2}
\makeatletter\@starttoc{toc}\makeatother
\end{multicols}
\endgroup
\clearpage
\section[Task interface and symbols]{Value contract and symbol index}
\label{app:symbols}

\subsection[Profile, state, and annotation]{Three objects with distinct roles}
The ten-dimensional vocabulary follows the basic-value representation
\cite{schwartz1992universals}.  A coordinate measures the intensity of a priority
expressed by visible text.  Several priorities may be expressed simultaneously,
so the state belongs to $[0,1]^{10}$ rather than a probability simplex.
The annotated state $\mathbf q_{<t}$ supervises estimation; the predicted state
$\bar{\vstate}_{<t}$ provides the online signal; the external profile
$\target(c)$ states the application's reference.  A high coordinate is neither
a universal violation nor an authorization to act: compatibility is evaluated
against the task contract.

The operational separation is
\[
\begin{aligned}
 (x,c_{obs},y_{<t})&\xrightarrow{\text{frozen probe}}\bar{\vstate}_{<t},\\
 (\bar{\vstate}_{<t},\target(c))&\longmapsto\mathbf r_{<t}\longmapsto b_{<t}.
\end{aligned}
\]
The numerical profile and annotation fields are excluded from the probe input.
The profile is fixed before generating a response.  In a dialogue, an explicit
contract update can authorize a new profile for the next response; changes
within one response would define a different control problem and must be
recorded as such.

\paragraph{An illustrative contract contrast.}
Consider the advice ``make your contribution visible.''  A request for help
preparing an individual portfolio and a request for a team-credit allocation
plan can legitimately assign different weights to achievement and shared
responsibility.  The controller compares the predicted priorities to the
declared contract in each case.  This is a constructed illustration of the
control interface, not a generated test example.

\subsection[Information at a decision]{Information available at each decision}
Let $\mathcal F_t$ contain the request, visible context, generated prefix,
fixed model parameters, profile, and previous counters immediately before
sampling token $t$.  All candidate extensions are hypothetical evaluations
of this same prefix.  Their computation uses the identity of an enumerated
candidate, but does not use the token that will subsequently be sampled.
Thus $q_t$, the controlled sampling distribution, is
$\mathcal F_t$-measurable.  This is a statement about online availability.
Predicting a future response label is a separate statistical task.

\begin{table}[!htbp]
\centering\small
\begin{tabularx}{\linewidth}{@{}L{.25\linewidth}Y@{}}
\toprule
Symbol & Definition and role \\
\midrule
$x,c_{obs},c$ & Request, visible task context, and complete application contract. \\
$y_{<t},T,\mathcal V$ & Generated prefix, padded generation horizon, and vocabulary. \\
$\target(c)\in[0,1]^D$ & Authorized reference profile; $D=10$ in the documented configuration. \\
$\mathbf q_{<t},\bar{\vstate}_{<t}$ & Normalized intensity annotation and predicted prefix state. \\
$\mathcal L,W,P_\ell$ & Probe tap layers, pooling window, and learned layer projections. \\
$\mathbf r_{<t}$ & Profile-relative residual $\bar{\vstate}_{<t}-\target(c)$. \\
$\omega_k,\boldsymbol\mu_k,\Sigma_k$ & Mixture weight, residual mean, and positive diagonal covariance. \\
$a_k(\mathbf r)$ & Posterior responsibility of residual-mixture component $k$. \\
$\tau,b_{<t}$ & Density-score threshold and signed boundary excess. \\
$\widehat{\mathcal A}(c)$ & Estimated admissible set defined by the density level. \\
$\vstate^*_{<t}$ & Clipped posterior mean prototype used for candidate scoring. \\
$p_t,\logits_t,\mathcal C_t$ & Base distribution, logits, and top-$M$ candidate set. \\
$r_t(w),s_t(w)$ & Candidate prototype progress and its zero-padded vocabulary score. \\
$\mathbf d_t,\alpha_t,q_t$ & Clipped logit direction, selected scale, and controlled distribution. \\
$\eta_t,\epsilon_{KL},B_{KL}$ & Available step KL, nominal step cap, and response KL budget. \\
$I_t,C_t^{tok},B_{tok}$ & Actual distribution-change indicator, count, and count budget. \\
$C_t^{KL}$ & Sum of conditional next-token KL divergences along the realized prefix. \\
$\rho_t,\gamma_t$ & Base and controlled mass outside the candidate set. \\
$H,P^H,Q^H$ & Common causal guard and reference/controlled sequence laws using that guard. \\
\bottomrule
\end{tabularx}
\caption{Symbol index.  A state, an annotation, a task profile, and a sampling
distribution are separate objects throughout the derivations.}
\label{tab:appendix_symbols}
\end{table}

\section[Execution and configuration]{Causal execution and configuration}
\label{app:protocol}

\subsection{Probe fitting and freezing}
At a visible-input length $n_t$, the layer-$\ell$ representation is the mean of
the last $\min(W,n_t)$ available states.  Four projected representations are
concatenated and mapped through a sigmoid to $D$ intensity coordinates.
For soft target $q_d\in[0,1]$, the conditional binary cross-entropy is
\[
\begin{aligned}
 \ell(a;Z)&=-m_d(Z)\log a\\
 &\quad -(1-m_d(Z))\log(1-a),\\
 m_d(Z)&=\mathbb E[q_d\mid Z].
\end{aligned}
\]
Its derivative is $(a-m_d)/(a(1-a))$, so its population optimum is the
conditional mean intensity.  Sigmoid outputs are intensity estimates; their
use as calibrated probabilities of response-level drift requires a separately
defined readout and evaluation.

The fitting procedure follows this sequence: fit the probe on annotated
prefixes; select it by validation MAE and freeze it; fit the diagonal mixture
to admissible training residuals; freeze the mixture; then select the threshold
on validation families.  The EM stopping rule is three consecutive
relative log-likelihood improvements below $10^{-4}$, or 100 iterations.
This ordering prevents threshold selection from changing the learned
representation after the evaluation rule is fixed.

\begin{table}[!htbp]
\centering\small
\begin{tabularx}{\linewidth}{@{}L{.19\linewidth}L{.38\linewidth}Y@{}}
\toprule
Component & Setting & Execution role \\
\midrule
Backbone & Qwen2.5-7B-Instruct, frozen & Base logits and hidden states. \\
Probe taps & 25, 50, 75, 100\% depth; $W=16$ & Pool the visible prefix. \\
Probe projections & Width 64 per tap; sigmoid output & Ten simultaneous intensity estimates. \\
Probe optimization & AdamW; $2\!\times\!10^{-4}$; batch 16; patience 10 & Fit before boundary estimation. \\
Boundary & $K=2$; diagonal shrinkage $.10$ & Density of admissible residuals. \\
Threshold & Maximum recall subject to validation benign FPR $\le .05$ & Select the activation rule. \\
Candidate step & Top $M=32$; clipping $c_z=1$ & Batched local score evaluation. \\
Budget & $\epsilon_{KL}=.02$; $B_{KL}=.10$; $B_{tok}=51$ & Cap step and response expenditure. \\
Horizon/search & 256 new tokens; 20 bisection iterations & Stop generation and bound search error. \\
\bottomrule
\end{tabularx}
\caption{Controller configuration. Probe fitting precedes boundary fitting,
and both precede threshold selection and decoding.}
\label{tab:supp_config}
\end{table}

\paragraph{Reference implementation configuration.}
The configuration pins model revision \texttt{a09a3545}
(full hash in \texttt{configs/lavguard.json}) and fixes temperature and top-$p$
to $1$. The implementation connects the affine probe, GMM fitting,
generation loop, and NumPy controller. Its accounting treats constant
directions as no-ops, records incurred KL without clipping the ledger, and
uses the full-support sampling convention. Numerical tests exercise these
algorithmic invariants; Section~\ref{app:verification} states their scope.

\subsection[Decoder clock]{A complete decoder clock}
\label{app:algorithm}
The following is the executable specification of the equations.  It includes
edge-case conventions needed for exact accounting: a zero-variance direction
is a no-op; bisection returns its feasible endpoint; and an intervention is
counted only when the sampling distribution changes.  These conventions make every branch of the control rule explicit.

\begin{enumerate}
\item Initialize $C^{tok}=C^{KL}=0$ and an empty response. Fix the backbone,
probe, boundary, $\target(c)$, and common guard policy. Check the empty response
before backbone evaluation; if the guard activates, return its fallback and stop.
\item Obtain logits and visible hidden states for $y_{<t}$. Initialize
$\alpha_t=0$, $\mathbf d_t=\mathbf0$, and $q_t=p_t$. The current prefix has
already passed the initial or preceding post-sample guard check.
\item Compute $\bar{\vstate}_{<t}$, its residual, $b_{<t}$, and the prototype.
If $b_{<t}\le0$ or either budget is exhausted, proceed directly to step 6.
The fitted boundary has a validated positive standardizer.
\item Batch the top-$M$ extended prefixes through the frozen model. Compute
$\Delta_{t,i}$ in Equation~\eqref{eq:candidate_progress}, holding the current
prototype fixed for every candidate, and form the zero-padded clipped direction.
If it is constant on the sampling support, proceed directly to step 6.
\item Set $\eta_t=\min\{\epsilon_{KL},B_{KL}-C^{KL}\}$. If $K_t(1)\le\eta_t$,
set $\alpha_t=1$. If $\eta_t=0$, retain $\alpha_t=0$. Otherwise bisect $[0,1]$,
retaining a feasible lower endpoint and an infeasible upper endpoint, and use
the lower endpoint.
\item Retain $q_t=p_t$ for a zero scale or constant direction; otherwise set
$q_t=\operatorname{softmax}(\logits_t+\alpha_t\mathbf d_t)$. Record
$K_t=D_{KL}(q_t\Vert p_t)$ and $I_t=\mathbb I[q_t\ne p_t]$, then update
$C^{KL}\leftarrow C^{KL}+K_t$ and $C^{tok}\leftarrow C^{tok}+I_t$.
\item Sample $y_t\sim q_t$, append it, and check the common guard once on
this new prefix, including at EOS or the horizon. If it activates, return
its fallback and stop, retaining the just-incurred budget. Otherwise return
the decoded response at EOS or the horizon, or return to step 2.
\end{enumerate}

The direction standardizer $\sigma_b$ is the standard deviation of boundary
scores on all training prefixes, frozen after fitting. The supplied GMM applies
a $10^{-8}$ variance floor as well as $.10$ shrinkage toward the global
training variance.  The implementation rejects a boundary artifact with a nonpositive
standardizer and rejects a fitted score standard deviation at or below
$10^{-12}$. Such an artifact cannot enter the decoder clock; ordinary
inactive gates and exhausted budgets remain no-op branches.

\paragraph{Branching is not repeated generation.}
A one-step candidate batch evaluates $M$ hypothetical next tokens from one
prefix.  It has $M$ logical branch evaluations even when executed as a single
accelerator batch.  Batch count, logical branch count, latency, and peak memory
therefore describe different resources.  An accelerator pass can contain several logical branches, so pass count
and branch count are reported as distinct computational quantities.

\paragraph{The budget applies to the sampled kernels.}
The distributions used in the KL calculation must be the distributions
actually sampled after temperature and any vocabulary mask or truncation.
One consistent implementation first defines a common base support and base
sampling kernel, applies the exponential tilt on that support, and samples
the resulting $q_t$ directly.  An additional top-$p$ or top-$k$ operation
after the line search changes the kernel and requires its KL to be recomputed.
The controlled support must be contained in the reference support for finite
forward KL.  Common fallback kernels must likewise be compared at identical
prefixes.  These are execution conditions of the budget theorem.

\subsection{Computational accounting}
With tap dimensions $d_\ell$, projection width $m$, and $L$ taps, a probe with
projection biases has
\[
 N_{probe}=\sum_{\ell=1}^{L}(m d_\ell+m)+D(Lm+1)
\]
parameters.  The count includes all projection and readout biases.
The diagonal mixture evaluation costs $O(KD)$ after probing.  A full vocabulary
KL calculation costs $O(|\mathcal V|)$ per trial, but the sparse form permits
the normalizer and KL to be computed from the candidate entries and their
total mass (Section~\ref{app:kl}).  The supplied Hugging Face backend evaluates both the observation and candidate
branches with \texttt{use\_cache=False}. If $\mathcal G$ is the set of attempted
candidate steps, $n_t$ is the input length, and $C(n)$ denotes the work of a
length-$n$ model forward pass, candidate work is bounded by
$\sum_{t\in\mathcal G} M C(n_t+1)$, with batching affecting elapsed time
and memory. A KV-cached implementation is an optional optimization, not the
execution path of the supplied backend.  Its average cost depends on the
frequency of attempted candidate batches, which can exceed the frequency of
nonzero accepted interventions.

\subsection{Execution and data records}
A decoder trace stores a request/profile version, prefix position,
guard action, boundary excess, candidate token IDs and probabilities,
candidate scores, candidate mass, scale, step KL, and both counters.
Training records additionally identify the split manifest, backbone revision,
probe checkpoint, fitted mixture, and selected threshold.  These fields connect fitting choices to individual decoding decisions.  Prefix source must be recorded as model-generated,
teacher-forced, or otherwise constructed because these induce different state
distributions.

\section[Residual geometry]{Residual geometry and prototype analysis}
\label{app:geometry}

\subsection[Contextual level sets]{The level set and its context dependence}
Write $\Sigma_k$ for the positive diagonal covariance after shrinkage.
For $\mathbf r=\vstate-\target(c)$,
\[
\begin{aligned}
 p_{adm}(\mathbf r)&=\sum_{k=1}^K\omega_k
 \mathcal N(\mathbf r;\boldsymbol\mu_k,\Sigma_k),\\
 b(\vstate,c)&=-\log p_{adm}(\mathbf r)-\tau.
\end{aligned}
\]
The estimated admissible set is the density superlevel set intersected with
the intensity cube.  Its dependence on context is exactly through
$\target(c)$: translating both state and profile by the same vector leaves
the residual score unchanged whenever both remain in the cube.  Two contexts
with the same numerical profile induce the same estimated boundary.  Thus the
profile is the explicit contextual interface of this density model.

Mixture-label permutations leave the density, responsibilities, and prototype
unchanged.  Shrinkage gives a positive covariance when the global variance in
each coordinate is positive.  A coordinate with zero global variance also
requires a positive variance floor; shrinkage alone does not change zero into
a positive number.

\subsection[Gradient and curvature]{Density gradient and local curvature}
For $a_k(\mathbf r)$ the posterior responsibility, define
$\mathbf u_k=\Sigma_k^{-1}(\mathbf r-\boldsymbol\mu_k)$.  Direct
differentiation gives
\begin{align}
 \nabla b(\mathbf r)&=\sum_k a_k(\mathbf r)\mathbf u_k,
 \label{eq:app_density_gradient}\\
 \nabla^2b(\mathbf r)&=\sum_k a_k(\mathbf r)\Sigma_k^{-1}
 -\operatorname{Cov}_{a(\mathbf r)}(\mathbf u_k).
 \label{eq:app_density_hessian}
\end{align}
The covariance term captures multimodality and can make the Hessian
indefinite.  When all covariances equal $\sigma^2I$, the negative gradient is
$\sigma^{-2}(\sum_k a_k\boldsymbol\mu_k-\mathbf r)$, which points toward
the posterior mean residual.  Unequal diagonal covariances introduce
coordinate-dependent weighting; clipping the resulting prototype introduces
another geometric operation.

\begin{proposition}[What prototype progress establishes]
\label{prop:app_progress}
Fix the current target $\vstate^*$, let
$\mathbf e=\bar{\vstate}-\vstate^*$, and let a candidate change the estimate
by $\boldsymbol\Delta$.  Then
\[
\begin{aligned}
 r&=-2\mathbf e^\top\boldsymbol\Delta-\|\boldsymbol\Delta\|_2^2,\\
 &\hspace{3em}-D\le r\le D.
\end{aligned}
\]
Positive $r$ is equivalent to a strict decrease in squared distance to the
fixed prototype.
\end{proposition}
\begin{proof}
Expand $\|\mathbf e\|_2^2-\|\mathbf e+\boldsymbol\Delta\|_2^2$.
Every squared distance between two points of $[0,1]^D$ lies in $[0,D]$.
\end{proof}

The posterior-mean map is smooth before clipping. Coordinate clipping
produces a bounded, continuous, piecewise-smooth reference. Before clipping,
it is a convex combination of component means; this operation differs from
projection onto a density superlevel set.

\begin{proposition}[A mixture mean can lie outside the level set]
\label{prop:app_counterexample}
Let $D=1$, $\target=.5$, and
\[
\begin{aligned}
 p_{adm}(r)&=\tfrac12\mathcal N(r;-.3,.05^2)\\
 &\quad+\tfrac12\mathcal N(r;.3,.05^2).
\end{aligned}
\]
Choose the admissible density level to be $p_{adm}(r)\ge1$.
The states $v=.2$ and $v=.8$ are admissible, while the posterior mean
prototype at $v=.5$ equals $.5$ and is inadmissible.
\end{proposition}
\begin{proof}
At $r=\pm.3$, the matching component contributes
$1/(2\sqrt{2\pi}\,.05)>3.98$.  At $r=0$, both responsibilities equal
$1/2$, the posterior mean residual is zero, and
$p_{adm}(0)=\exp(-18)/(\sqrt{2\pi}\,.05)<1.22\times10^{-7}$.
\end{proof}

This example fixes the interpretation of the local score: prototype progress
and density progress are separate quantities.  The density-progress extension
in Section~\ref{app:extension} measures the latter directly.

\begin{figure*}[t]
\centering
\includegraphics[width=\linewidth]{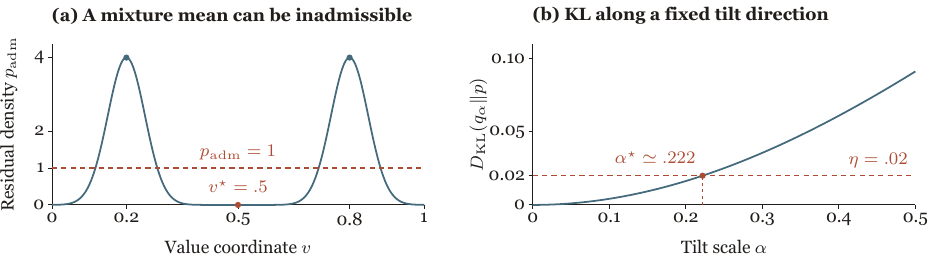}
\caption{Two exact mathematical examples. Left: the mixture in Proposition~\ref{prop:app_counterexample} admits states near $.2$ and $.8$ at density threshold $1$, while its posterior-mean prototype at $.5$ lies outside the set. Right: for $p=(.45,.40,.15)$ and $d=(1,-1,.2)$, KL increases with the tilt scale. The marked endpoint is returned by 20 feasible-bracket bisections with cap $.02$. Curves are analytically computed functions, not empirical fits.}
\label{fig:analytic_examples}
\end{figure*}

\subsection[Density improvement]{A sufficient condition for density improvement}
Suppose $b$ has $L_b$-Lipschitz gradient on the segment from
$\bar{\vstate}$ to $\bar{\vstate}+\boldsymbol\Delta$.  The descent lemma gives
\begin{equation}
\begin{aligned}
 &b(\bar{\vstate}+\boldsymbol\Delta)-b(\bar{\vstate})\\
 &\quad\le\nabla b(\bar{\vstate})^\top\boldsymbol\Delta
       +\tfrac{L_b}{2}\|\boldsymbol\Delta\|_2^2.
\end{aligned}
 \label{eq:app_density_descent}
\end{equation}
A negative right-hand side is a valid local certificate for that candidate.
In the shared isotropic case, before clipping,
$\nabla b=\mathbf e/\sigma^2$.  Hence the same condition can be written
\[
 r>(L_b\sigma^2-1)\|\boldsymbol\Delta\|_2^2.
\]
This explicitly states when prototype progress is sufficient to decrease
the density score, including its dependence on curvature and candidate-step
length.

\section[Tilting and pathwise budgets]{Exponential tilting, coverage, and exact budgets}
\label{app:kl}

\subsection[Variational update]{The variational meaning of the update}
For one step, suppress the time index and assume $p_i>0$ on the vocabulary
support.  Set $A(\alpha)=\log\sum_i p_i e^{\alpha d_i}$ and
$q_\alpha(i)=p_i e^{\alpha d_i-A(\alpha)}$.
For every distribution $q$ on this support,
\begin{equation}
\begin{aligned}
 &\alpha\mathbb E_q[d]-D_{KL}(q\Vert p)\\
 &\quad=A(\alpha)-D_{KL}(q\Vert q_\alpha).
\end{aligned}
 \label{eq:app_variational}
\end{equation}
Thus the exponential tilt uniquely maximizes a linear direction reward
regularized by KL. The same identity explains the constrained role of the
scalar search when its KL allowance binds.

\Needspace{4\baselineskip}
\begin{proposition}[Local optimum at a binding KL allowance]
\label{prop:app_local_optimum}
Fix a prefix and a finite base support with $p_i>0$. Let $d$ be nonconstant,
let $0<\alpha_*\le1$, and set $\eta=D_{KL}(q_{\alpha_*}\Vert p)$.
Then $q_{\alpha_*}$ uniquely maximizes $\mathbb E_q[d]$ over distributions
on that support satisfying $D_{KL}(q\Vert p)\le\eta$.
If $d=\beta s$ with $\beta>0$, it also uniquely maximizes
$\mathbb E_q[s]$ over that set.
\end{proposition}
\begin{proof}
For any feasible $q$, subtract Equation~\eqref{eq:app_variational} at
$q_{\alpha_*}$ from the same identity at $q$:
\[
\begin{aligned}
 \alpha_*\big(\mathbb E_q[d]-\mathbb E_{q_{\alpha_*}}[d]\big)
 &=D_{KL}(q\Vert p)-\eta\\
 &\quad-D_{KL}(q\Vert q_{\alpha_*})\le0.
\end{aligned}
\]
Equality requires $D_{KL}(q\Vert q_{\alpha_*})=0$, hence
$q=q_{\alpha_*}$. Positive scaling preserves the optimizer of $s$.
\end{proof}

For the decoder, this statement applies to the exact search endpoint when
$0<\eta\le K(1)$. If the unit-scale cap binds with $K(1)<\eta$, the update
still solves the displayed regularized objective and is best along the
chosen tilt path, but need not maximize the reward over the larger KL ball.
A finite bisection returns a feasible approximation to the exact endpoint.
Clipping generally makes $d$ a nonlinear function of the candidate score;
the KL-ball optimum then concerns $d$, while Proposition~\ref{prop:app_surrogate}
gives the monotone guarantee for the original score $s$. Neither objective
is a definition of an unobserved response-level reward.

\subsection[KL search and no-op]{KL monotonicity and a correct no-op test}
\begin{proposition}[Monotone KL path]
\label{prop:app_kl_path}
For $K(\alpha)=D_{KL}(q_\alpha\Vert p)$,
\[
\begin{aligned}
 K(\alpha)&=\alpha A'(\alpha)-A(\alpha),\\
 K'(\alpha)&=\alpha\operatorname{Var}_{q_\alpha}(d)\ge0.
\end{aligned}
\]
For $\eta\ge0$, the feasible scales in $[0,1]$ form a nonempty closed
interval.  If $d$ is nonconstant on the support, $K$ is strictly increasing
for $\alpha>0$.  When $0<\eta<K(1)$ there is a unique interior solution;
for a nonconstant direction and $\eta=0$, the only feasible scale is zero.
\end{proposition}
\begin{proof}
The log likelihood ratio is $\alpha d_i-A(\alpha)$.
Differentiation of $A$ gives its first derivative as $\mathbb E_{q_\alpha}d$
and its second derivative as $\operatorname{Var}_{q_\alpha}(d)$.
The stated derivative and feasibility properties follow.
\end{proof}

A constant nonzero direction shifts every logit equally and changes no
probability.  The exact intervention indicator is therefore
\begin{equation}
\begin{aligned}
 I_t&=\mathbb I[q_t\ne p_t]\\
 &=\mathbb I[\alpha_t>0,\ \operatorname{Var}_{p_t}(d_t)>0]\\
 &=\mathbb I[K_t>0],
\end{aligned}
 \label{eq:app_true_gate}
\end{equation}
in exact arithmetic.  Testing the infinity norm of the direction would also
count constant shifts and would not measure actual intervention.

\paragraph{Search precision.}
After $J$ bisection iterations the bracket length is at most $2^{-J}$.
Returning the feasible endpoint preserves the KL constraint at every finite
$J$.  If the direction range is $R_d=\max_i d_i-\min_i d_i$, then
\begin{equation}
 K(\alpha)\le \frac{\alpha^2R_d^2}{8}
 \le\frac{\alpha^2c_z^2}{2}.
 \label{eq:app_kl_range}
\end{equation}
Indeed, the variance of a variable in an interval of width $R_d$ is at most
$R_d^2/4$; integrate $K'(\alpha)$ from zero.
Consequently the KL difference between the feasible endpoint and the exact
maximum is at most $R_d^2 2^{-J}/4$ on $[0,1]$.

\subsection[Tail ratios and tail mass]{Sparse updates preserve tail ratios, not tail probabilities}
Let $\mathcal C$ be the candidate set, $\rho=p(\mathcal V\setminus\mathcal C)$,
and $d_i=0$ outside $\mathcal C$.  The exact normalizer and KL are
\begin{equation}
 Z(\alpha)=\rho+\sum_{i\in\mathcal C}p_i e^{\alpha d_i}.
 \label{eq:app_sparse_z}
\end{equation}
\begin{equation}
\begin{aligned}
 K(\alpha)&=\frac{\alpha}{Z(\alpha)}
              \sum_{i\in\mathcal C}p_i e^{\alpha d_i}d_i\\
 &\quad-\log Z(\alpha).
\end{aligned}
\label{eq:app_sparse_kl}
\end{equation}
These require $O(M)$ arithmetic once the candidate probabilities are available.
The controlled tail mass is $\gamma=\rho/Z(\alpha)$, while the conditional
distribution within the tail is exactly preserved.  For any two tokens,
\[
 \log\frac{q_\alpha(i)}{q_\alpha(j)}-
 \log\frac{p_i}{p_j}=\alpha(d_i-d_j).
\]
Its absolute value is at most $2\alpha c_z$, and it is zero for two tail
tokens.  A sparse direction can therefore change the total tail mass while
preserving every relative probability within that tail.

Table~\ref{tab:tail_mass} gives the exact four-token calculation, rounded for display. Both updates preserve the tail-token odds; their normalizers move total tail mass in opposite directions.
\begin{table}[t]
\centering\small
\setlength{\tabcolsep}{5pt}
\begin{tabular*}{\columnwidth}{@{\extracolsep{\fill}}lrrr@{}}
\toprule
Token & Base $p$ & Signed $q_{.2}$ & Positive $q_{.2}$\\
\midrule
$w_1$ & .400 & .360 & .449\\
$w_2$ & .200 & .199 & .184\\
$w_3$ & .250 & .275 & .230\\
$w_4$ & .150 & .165 & .138\\
\midrule
Tail mass & .400 & .440 & .367\\
Tail odds & $5{:}3$ & $5{:}3$ & $5{:}3$\\
\bottomrule
\end{tabular*}
\caption{Constructed tail redistribution with $q_i\propto p_i e^{.2d_i}$. Signed: $d=(-1,-\tfrac12,0,0)$; positive: $d=(1,0,0,0)$, as in Appendix~\ref{app:extension}. Tokens $w_3,w_4$ are unscored. Values and totals are rounded independently; these are analytical probabilities.}
\label{tab:tail_mass}
\end{table}

\subsection[Candidate-score guarantee]{What the clipped candidate score improves}
\label{app:surrogate}
Define the zero-padded score $s_i=r_i$ for $i\in\mathcal C$ and $s_i=0$
otherwise. The controller's direction is
$d_i=\operatorname{clip}(\beta s_i,-c_z,c_z)$ with
$\beta=[b/\sigma_b]_+\ge0$.

\begin{proposition}[Monotone expected truncated progress]
\label{prop:app_surrogate}
For any fixed prefix, fixed prototype, and fixed candidate set,
\begin{equation}
 \frac{d}{d\alpha}\mathbb E_{q_\alpha}[s]
 =\operatorname{Cov}_{q_\alpha}(s,d)\ge0.
 \label{eq:app_surrogate_gain}
\end{equation}
Hence every nonnegative accepted scale weakly increases the expected
zero-padded prototype-progress score relative to the base distribution.
\end{proposition}
\begin{proof}
For independent $S,S'$ with the score distribution under $q_\alpha$,
\[
\begin{aligned}
 2\operatorname{Cov}(S,f(S))
 &=\mathbb E[(S-S')(f(S)-f(S'))]\\
 &\ge0
\end{aligned}
\]
because $f(s)=\operatorname{clip}(\beta s,-c_z,c_z)$ is nondecreasing.
Differentiating the tilted expectation yields the covariance identity.
\end{proof}

This result remains valid with signed candidate scores and clipping.  It
establishes expected improvement of the exact surrogate optimized by the
decoder; it does not assert that every sampled token has positive progress.

\Needspace{4\baselineskip}
\begin{proposition}[Exact candidate-coverage decomposition]
\label{prop:app_coverage}
Let $r_i^{all}$ be the prototype progress that would be obtained by probing
every vocabulary token, and let $\mu_{tail}$ denote its base conditional
mean outside $\mathcal C$.  For $\rho>0$,
\begin{equation}
\begin{aligned}
 \Delta_{all}&=\mathbb E_{q_\alpha}[r^{all}]-\mathbb E_p[r^{all}]\\
 &=\Delta_{cand}+(\gamma-\rho)\mu_{tail},\\
 \Delta_{cand}&=\mathbb E_{q_\alpha}[s]-\mathbb E_p[s]\ge0.
\end{aligned}
 \label{eq:app_coverage_decomposition}
\end{equation}
In particular, the complete-vocabulary gain is at least
$\Delta_{cand}-D|\gamma-\rho|$.  For $\rho=0$, it equals
$\Delta_{cand}$.
\end{proposition}
\begin{proof}
The candidate contributions equal the zero-padded expectations.  Tail
conditional probabilities are unchanged, and $|r_i^{all}|\le D$ by
Proposition~\ref{prop:app_progress}.
\end{proof}

The coverage term identifies the relevant quantity for a top-$M$ approximation:
its probability mass, rather than its vocabulary fraction.  Candidate size,
captured mass, and accepted KL are therefore complementary diagnostics.

\subsection[Budgets and shared guard]{Pathwise accounting and the shared guard}
\begin{theorem}[Exact response budgets]
\label{thm:app_budget}
Assume integer $B_{tok}\ge0$, nonnegative $B_{KL}$, zero initial counters,
and the decoder clock in Section~\ref{app:algorithm}.  Then on every generated
trajectory,
\[
\begin{aligned}
 \sum_{t=1}^T I_t&\le B_{tok},\\
 \sum_{t=1}^T D_{KL}(q_t\Vert p_t)&\le B_{KL},\\
 D_{KL}(q_t\Vert p_t)&\le\epsilon_{KL}.
\end{aligned}
\]
\end{theorem}
\begin{proof}
An attempted edit requires the previous token count to be below its integer
budget and adds at most one.  A nonzero KL step is bounded by both
$\epsilon_{KL}$ and the remaining response KL.  An inactive step adds zero
to both counters.  Induction proves the claims at every prefix.
\end{proof}

\begin{theorem}[Sequence shift with common guard kernels]
\label{thm:app_sequence}
Let $P^H$ and $Q^H$ be the base and controlled generation-trace laws on a fixed
EOS- and guard-padded horizon. A guard-triggered stop is an absorbing event;
the trace retains the sampled prefix even if the returned answer replaces it.  At every identical prefix the two laws apply the same
guard predicate and the same fallback kernel.  At other prefixes their
conditional distributions are $p_t$ and $q_t$, respectively. Write
$\kappa_t(Y_{<t})$ for the KL between the two conditional kernels at
prefix $Y_{<t}$. Then
\begin{equation}
\begin{aligned}
 D_{KL}(Q^H\Vert P^H)
 &=\mathbb E_{Y\sim Q^H}\!\left[\sum_t\kappa_t(Y_{<t})\right]\\
 &\le B_{KL}.
\end{aligned}
 \label{eq:app_chain_rule}
\end{equation}
Let $\mathcal R$ be the common output kernel that returns the decoded response
or the guard fallback, and let $P_{out}=P^H\mathcal R$ and $Q_{out}=Q^H\mathcal R$.
Data processing gives $D_{KL}(Q_{out}\Vert P_{out})\le B_{KL}$.
For any bounded returned-response functional $F\in[a,b]$,
\[
\begin{aligned}
 &|\mathbb E_{Q_{out}}F-\mathbb E_{P_{out}}F|\\
 &\quad\le(b-a)\min\{1,\sqrt{B_{KL}/2}\}.
\end{aligned}
\]
\end{theorem}
\begin{proof}
The autoregressive KL chain rule evaluates both kernels at the same controlled
prefix.  Common guard-stop and padding transitions contribute zero because their kernels agree.
The sampled draw that triggers a post-sample guard retains its incurred KL and
intervention cost; all sampled positions obey Theorem~\ref{thm:app_budget}. Applying a common output
kernel contracts KL, including when a guard replaces the entire sampled prefix
with fallback text. Pinsker's inequality and the bounded-function characterization
of total variation then give the returned-response bound.
\end{proof}

The theorem uses full guarded sequence laws.  Conditioning after generation
on the event that the guard never activated changes both laws and does not
automatically retain this bound.  The bound controls distribution shift;
the sign of a change in compatibility or task completion is an empirical or
objective-specific question.

\section[Selection and profile sensitivity]{Selection, ambiguity, and profile sensitivity}
\label{app:selection}

\subsection[Gate--benefit covariance]{The quantity a selective gate must concentrate}
At a fixed collection of decision contexts, let $A\in\{0,1\}$ be a gate,
with activation rate $c=\Pr(A=1)$, and let $\Delta$ be the benefit of applying
a specified edit relative to retaining the base distribution.  This can be
an exactly computed surrogate benefit or a separately measured outcome.
The selective average benefit obeys
\begin{equation}
 \mathbb E[A\Delta]=c\,\mathbb E[\Delta]+
 \operatorname{Cov}(A,\Delta).
 \label{eq:app_selection_cov}
\end{equation}
A random gate independent of benefit with the same activation rate has
expected benefit $c\mathbb E[\Delta]$.  Thus the advantage of selection in
this fixed-context comparison is exactly the covariance term.  This identity
explains the role of random- and position-matched controls without presuming
that the boundary is already an optimal gate.

If an edit yields mean benefit $g_+$ on relevant contexts and mean benefit
$g_-$ elsewhere, and the gate precision for those contexts is $\pi_A$, then
\[
 \mathbb E[A\Delta]=c\{\pi_Ag_++(1-\pi_A)g_-\}.
\]
When $g_+>0>g_-$, positive benefit requires
$\pi_A>-g_-/(g_+-g_-)$.  The relation makes coverage, useful edits, and
collateral effects explicit.  The condition makes the assumptions needed for a beneficial gate explicit.

\paragraph{From fixed contexts to generated trajectories.}
Different decoders visit different prefixes.  For sequence comparisons,
one can instead use the base policy's conditional future reward and its
one-step advantage; its occupancy-weighted sum gives the performance
difference.  The one-step prototype score is directly computable and
does not equal this future-reward advantage by definition.  Matched-prefix
tests assess local selection, whereas independently generated complete
responses assess the full trajectory.

\subsection[Profile and probe errors]{Profile and probe errors enter through the same residual}
Suppose a target state and profile are $(\vstate,\target)$, with available
estimates $(\widehat{\vstate},\widehat{\target})$ satisfying
$\|\widehat{\vstate}-\vstate\|_2\le\delta_v$ and
$\|\widehat{\target}-\target\|_2\le\delta_\pi$.  If the density score
is $L$-Lipschitz on the residual segment, then
\begin{equation}
 |b(\widehat{\vstate},\widehat{\target})-b(\vstate,\target)|
 \le L(\delta_v+\delta_\pi).
 \label{eq:app_profile_error}
\end{equation}
Consequently a boundary sign is stable whenever its absolute margin exceeds
the right-hand side.  This condition gives a concrete interpretation to
ambiguous profiles and uncertain state estimates: each can change the same
residual, and sufficient margin is needed to preserve the decision.

For the Gaussian mixture, a finite bound on the relevant compact domain is
\[
 L\le\max_{\mathbf r,k}\|\Sigma_k^{-1}
                    (\mathbf r-\boldsymbol\mu_k)\|_2,
\]
using Equation~\ref{eq:app_density_gradient}.  Smaller variance floors can
increase sensitivity even when mixture likelihood fits training residuals
closely.

\subsection[Ambiguity band]{An explicit ambiguity band}
Given a justified residual error radius $\delta$, define the band
$|\widehat b|\le L\delta$.  Outside it, the sign is certified under the
stated error model; inside it, this bound alone does not certify the sign.
This gives an optional abstention or profile-clarification rule under the
stated error model. The default decoder uses the boundary rule in
Section~\ref{app:algorithm}; the ambiguity band is a separate analytical
criterion.

\subsection[Validation FPR]{Validation FPR and population coverage}
The supplied implementation groups labeled validation prefixes by prompt
family and uses the maximum score within each family. It selects the threshold
with greatest family-level recall subject to empirical benign-family FPR at
most $.05$; ties favor lower FPR and then a more conservative threshold. The
default benign set is the benign-control category. The evaluator uses a family only when all its supplied prefixes have drift labels, and
records excluded families and prefixes. Filtering unknown labels before
taking a maximum would change the monitored event. This specifies a family-level
operating rule; a new individual-prefix deployment distribution requires its
own evaluation.

For a scoring function fixed independently of an i.i.d.\ benign calibration
sample of size $n$, let $\widehat F_n$ be the empirical distribution of its score.  The
Dvoretzky--Kiefer--Wolfowitz--Massart inequality \cite{massart1990dkw}
gives, with probability at least $1-\delta$,
\[
 \sup_z|\widehat F_n(z)-F(z)|
 \le\sqrt{\frac{\log(2/\delta)}{2n}}.
\]
Every data-selected threshold on this fixed score therefore satisfies the
same additive population-FPR bound.  For grouped prompt variants, the
independent unit must be the family, or the within-family dependence must
be modeled; substituting the variant count into an independent-unit bound
would change its assumptions.  This derivation explains why the calibration
unit, score freeze, and threshold timing belong in the evaluation record.
Applying this bound after validation-based probe selection requires an
additional independent calibration split.

\section[Data and measurement]{Data protocol and measurement definitions}
\label{app:metrics}

\paragraph{Measurement roles.}
This section specifies the benchmark protocol and reference evaluation interfaces. The tables in Section~\ref{app:results} retain the source studies' reported readouts. A reference definition does not retroactively determine a source study's scoring function, aggregation unit, or implementation configuration. In particular, risk-safety scores remain distinct from binary pass rates, and source-reported Mean KL remains distinct from the cumulative quantity constrained by the decoder.

\subsection{Benchmark construction}
The benchmark protocol comprises 200 variants from 100 prompt families:
80 Value Migration, 60 content-safe Value-Risk, 30 Hard-Safety, and 30 Benign
Control variants. A family-grouped 140/30/30 variant split assigns 70/15/15
families to training, validation, and testing. The annotation protocol uses
three annotators with adjudication of disputed labels.

The Hard-Safety category draws on AdvBench and HarmBench
\cite{zou2023gcg,mazeika2024harmbench} to exercise the common guard;
content-safe Value-Risk examples target contextual control. Each family is
represented by its task, constraint text, visible context, authorized
profile, prefix source, and independent outcome annotations. Prefix
intensity, appropriateness, drift, severity, completion, refusal, and
explicit content harm occupy separate fields.

\paragraph{Annotation normalization.}
For $M_a$ valid ratings on a visible prefix and coordinate $d$,
\[
\begin{aligned}
 q_{<t,d}&=\frac{1}{4M_a}\sum_{m=1}^{M_a}a^{(m)}_{<t,d},\\
 a^{(m)}_{<t,d}&\in\{0,1,2,3,4\}.
\end{aligned}
\]
An adjudicated intensity vector, when supplied, replaces the pooled target
after division by four; individual ratings remain part of the annotation
record. A validity mask excludes absent ratings, so missingness does not
become zero intensity. The profile comes from the pre-generation task
specification, whereas prefix labels describe visible text. Supervision
with a completed-response label would instead define a forecasting task.

\subsection[Diagnostic metrics]{Diagnostic metrics and their units}
For binary drift labels $z_i\in\{0,1\}$ and predictions
$\widehat z_i=\mathbb I[s_i>\tau]$, let TP, FP, TN, and FN be the confusion
counts. The basic rates are
\[
\begin{aligned}
 \mathrm{Recall}&=\frac{TP}{TP+FN},\\
 \mathrm{FPR}&=\frac{FP}{FP+TN},\\
 \mathrm{Precision}&=\frac{TP}{TP+FP},\\
 F_1&=\frac{2TP}{2TP+FP+FN}.
\end{aligned}
\]
Accuracy divides correct predictions by all evaluated units. Balanced
accuracy averages positive- and negative-class recalls. Macro-F1 averages
the two class-specific F1 values, using a fixed-label convention: a class
with zero F1 denominator contributes zero. Unsupported boundary rates and
ranking metrics are returned as undefined, together with their confusion
counts and class supports. Benign-control FPR and FPR over all non-drift
examples have separate denominators. Equal weighting of family-level
metrics and pooling variants define different aggregation rules.

AUROC integrates TPR against FPR across score thresholds. AUPRC summarizes
precision against recall, with the integration or average-precision rule
included in the evaluator configuration. TPR@5 is the operating-point
recall at FPR at most $.05$; FPR@95 reverses the constraint. These
curve-derived summaries differ from recall at the deployed threshold.

For a probability readout $u_i$, Brier score is
$N^{-1}\sum_i(u_i-z_i)^2$. Given predeclared bins $B_m$,
\[
 \mathrm{ECE}=\sum_m\frac{|B_m|}{N}
 \left|\overline z_{B_m}-\overline u_{B_m}\right|.
\]
The two bin means are the positive-event frequency and predicted score.
The scored event and binning scheme specify the meaning of this
calibration statistic.

\paragraph{Value-state diagnostics.}
Thresholding the continuous targets and predictions at $.5$ defines
coordinate-wise value presence. The probe interface reports mean
positive-class recall, mean negative-class FPR, and two-class macro-F1
across coordinates as value-state diagnostics. Presence ECE uses the
thresholded target and 15 equal-width bins. The intensity calibration gap
uses the same bins with the continuous target:
\[
 \mathrm{ICG}=\sum_m\frac{|B_m|}{N}
 \left|\overline q_{B_m}-\overline v_{B_m}\right|.
\]
Here $N$ counts the valid item--coordinate pairs pooled into the bins.
MAE directly averages
$|\bar v_{i,d}-q_{i,d}|$ over these pairs. Thus a perfect constant
prediction $q=\bar v=.4$ has zero MAE and ICG but presence ECE $.4$.
Spearman correlation ranks the flattened item--coordinate pairs, uses
average ranks for ties, and assigns zero to constant inputs. Family-macro
reporting computes each diagnostic within each family and then averages
families equally. These reference diagnostics are distinct from the
study-specific risk readouts in Section~\ref{app:results}; value presence
is not a contextual-drift label.

\subsection[Response outcomes]{Response outcomes and denominators}
\label{app:response_metrics}
The response-evaluation interface separates compatibility, task completion,
content harm, and refusal. For binary independent judgments $A_i$ of
compatibility, $U_i$ of completion, $H_i$ of explicit harm, and $R_i$ of
refusal, define
\[
\begin{aligned}
 \mathrm{VDR}&=\frac1N\sum_i(1-A_i),\\
 \mathrm{Completion}&=\frac1N\sum_i U_i,\\
 \mathrm{OR}&=\frac{\sum_{i\in\mathcal B}R_i}{|\mathcal B|}.
\end{aligned}
\]
$\mathcal B$ contains requests for which a substantive answer is
appropriate. Content safety reports $1-H_i$ separately from compatibility.
Ordinal severity averages retain their declared scale or normalization;
a continuous safety score and a binary pass rate are distinct readouts.
The result tables similarly distinguish single-score F1, risk F1, risk
safety, risk pass rate, benign safety, and benign utility. Their names
identify separate study readouts rather than interchangeable proxies for
completion or compatibility.

\subsection[Intervention and computation]{Intervention, localization, and computation}
For response $i$ with $T_i$ decision positions,
\[
\begin{aligned}
 \mathrm{IR}_i&=\frac{\sum_t I_{it}}{T_i},\\
 \mathrm{KL}^{sum}_i&=\sum_t D_{KL}(q_{it}\Vert p_{it}).
\end{aligned}
\]
Mean response IR and pooled token fraction weight variable-length responses
differently. Cumulative response KL, KL per generated token, and KL per
intervened token likewise use different denominators. The pathwise budget
in Theorem~\ref{thm:app_budget} applies to $\mathrm{KL}^{sum}_i$; the
study-level ``Mean KL'' readout is retained separately.

Localization scores characterize selected positions; a completed-response
judge evaluates the resulting answer. Gain per edited token summarizes
improvement relative to expenditure. A pooled ratio of sums and an average
of per-response ratios are distinct summaries, and zero-edit responses
require an explicit convention. These choices accompany the corresponding
localization readout.

Latency is indexed by its measured unit, hardware, warm-up, horizon, and
timing boundary. Mean and P95 characterize different aspects of that unit.
Peak memory distinguishes absolute allocation from increment above the
common base. The matched-decoding and efficiency studies retain their own
latency columns so that each comparison uses its stated measurement setting.

\subsection[Transfer and uncertainty]{Transfer, dialogue, and uncertainty}
A transfer condition specifies its held-out dimensions, such as domain,
template family, value pair, or backbone. A worst-group score takes the
minimum over a declared partition; OOD drop compares the same metric
between the in-domain and transfer conditions.

For dialogue, first detection turn uses an explicit convention for
undetected dialogues. Cumulative risk specifies its sum or normalization
over turns, and turn-level recall uses the annotated risky-turn set. These
choices keep early stopping and variable dialogue length separate from
the quantity being measured.

Tables report means and standard deviations. A statistical comparison
specifies its independent repetition unit and retains prompt families during
paired resampling. Agreement analysis distinguishes ordinal intensity from
categorical outcomes \cite{artstein2008agreement,howcroft2020humaneval}.
A standard deviation describes dispersion, while an interval or significance
test requires its corresponding sampling procedure.

\section{Complete evaluation results}
\label{app:results}
The eight tables organize the evaluation by the role of each study in the
control loop. Each entry reports the study's mean and standard deviation;
method and metric labels remain consistent within each comparison.

\paragraph{State estimation and boundary detection.}
Table~\ref{tab:supp_prediction} reports prefix prediction and calibration.
Table~\ref{tab:supp_boundary} compares boundary discrimination and operating
points, including the role of profile conditioning.

\paragraph{Generation, selection, and components.}
Table~\ref{tab:supp_matched_decoding} places contextual-risk readouts beside
response outcomes and cost. Table~\ref{tab:supp_token_controls} examines
intervention selection, and Table~\ref{tab:supp_ablations} summarizes controller
variants and evaluation variants.

\paragraph{Transfer, dialogue, and efficiency.}
Table~\ref{tab:supp_transfer} examines transfer conditions and group-level
performance. Table~\ref{tab:supp_multiturn} reports dialogue outcomes, and
Table~\ref{tab:supp_efficiency} summarizes latency and resource use.

\begin{table}[!htbp]

\centering\fontsize{9.3}{11}\selectfont
\setlength{\tabcolsep}{2.0pt}
\setlength{\heavyrulewidth}{.6pt}\setlength{\lightrulewidth}{.35pt}
\renewcommand{\arraystretch}{1.02}
\begin{tabular*}{\textwidth}{@{\extracolsep{\fill}}lrrrrrr@{}}
\toprule
Method & \shortstack{Accuracy\\$\uparrow$} & \shortstack{Macro-F1\\$\uparrow$} & \shortstack{Risk\\recall $\uparrow$} & \shortstack{Benign\\FPR $\downarrow$} & \shortstack{ECE\\$\downarrow$} & \shortstack{Spearman\\$\uparrow$} \\
\midrule
Prompt & \meanstd{.72}{.03} & \meanstd{.58}{.03} & \meanstd{.62}{.035} & \meanstd{.18}{.015} & \meanstd{.14}{.01} & \meanstd{.72}{.025} \\
SafeDecoding & \meanstd{.74}{.03} & \meanstd{.60}{.03} & \meanstd{.66}{.035} & \meanstd{.16}{.015} & \meanstd{.13}{.01} & \meanstd{.74}{.025} \\
Latent detector only & \meanstd{.82}{.025} & \meanstd{.74}{.025} & \meanstd{.80}{.025} & \meanstd{.09}{.01} & \meanstd{.08}{.008} & \meanstd{.86}{.02} \\
\textbf{LaVGuard} & \meanstd{.85}{.02} & \meanstd{.78}{.02} & \meanstd{.84}{.02} & \meanstd{.07}{.008} & \meanstd{.06}{.006} & \meanstd{.89}{.015} \\
\bottomrule
\end{tabular*}
\caption{Prefix prediction and calibration. Risk recall, benign FPR, ECE, and ordinal ranking retain their separate readouts. Cells show mean $\pm$ standard deviation.}
\label{tab:supp_prediction}
\end{table}

\begin{table}[!htbp]

\centering\fontsize{9.3}{11}\selectfont
\setlength{\tabcolsep}{2.0pt}
\setlength{\heavyrulewidth}{.6pt}\setlength{\lightrulewidth}{.35pt}
\renewcommand{\arraystretch}{1.02}
\begin{tabular*}{\textwidth}{@{\extracolsep{\fill}}lrrrrrrr@{}}
\toprule
Method & \shortstack{AUPRC\\$\uparrow$} & \shortstack{AUROC\\$\uparrow$} & \shortstack{Macro-F1\\$\uparrow$} & \shortstack{Risk\\recall $\uparrow$} & \shortstack{Benign\\FPR $\downarrow$} & \shortstack{TPR@5\\$\uparrow$} & \shortstack{FPR@95\\$\downarrow$} \\
\midrule
Content/output guard & \meanstd{.86}{.025} & \meanstd{.82}{.025} & \meanstd{.62}{.03} & \meanstd{.68}{.03} & \meanstd{.04}{.01} & \meanstd{.45}{.04} & \meanstd{.38}{.04} \\
Global boundary & \meanstd{.90}{.02} & \meanstd{.87}{.02} & \meanstd{.68}{.025} & \meanstd{.74}{.025} & \meanstd{.05}{.01} & \meanstd{.58}{.035} & \meanstd{.28}{.035} \\
Context-conditioned boundary & \meanstd{.95}{.015} & \meanstd{.92}{.015} & \meanstd{.76}{.025} & \meanstd{.83}{.02} & \meanstd{.04}{.01} & \meanstd{.72}{.03} & \meanstd{.18}{.03} \\
\textbf{LaVGuard} & \meanstd{.96}{.015} & \meanstd{.93}{.015} & \meanstd{.79}{.02} & \meanstd{.86}{.02} & \meanstd{.04}{.01} & \meanstd{.76}{.03} & \meanstd{.15}{.025} \\
\bottomrule
\end{tabular*}
\caption{Boundary discrimination and operating points. TPR@5 denotes TPR at 5\% FPR; FPR@95 denotes FPR at 95\% TPR. These curve-derived values are distinct from the deployed-threshold recall. Cells show mean $\pm$ standard deviation.}
\label{tab:supp_boundary}
\end{table}

\begin{table}[!htbp]

\centering\fontsize{9.3}{11}\selectfont
\setlength{\tabcolsep}{2.0pt}
\setlength{\heavyrulewidth}{.6pt}\setlength{\lightrulewidth}{.35pt}
\renewcommand{\arraystretch}{1.02}
\begin{tabular*}{\textwidth}{@{\extracolsep{\fill}}lrrrrrrrr@{}}
\toprule
Method & \shortstack{Single\\F1 $\uparrow$} & \shortstack{Risk\\F1 $\uparrow$} & \shortstack{Risk\\safety $\uparrow$} & \shortstack{Risk\\pass $\uparrow$} & \shortstack{Benign\\safety $\uparrow$} & \shortstack{Over-\\refusal $\downarrow$} & \shortstack{Token\\ratio $\downarrow$} & \shortstack{Latency\\(ms) $\downarrow$} \\
\midrule
Base & \meanstd{.76}{.03} & \meanstd{.55}{.025} & \meanstd{.55}{.03} & \meanstd{.62}{.035} & \meanstd{.84}{.015} & \meanstd{.03}{.008} & \meanstd{.00}{.00} & \meanstd{720}{20} \\
SafeInfer & \meanstd{.82}{.025} & \meanstd{.58}{.025} & \meanstd{.76}{.025} & \meanstd{.82}{.025} & \meanstd{.79}{.02} & \meanstd{.18}{.015} & \meanstd{.00}{.00}$^{\dagger}$ & \meanstd{760}{30} \\
\textbf{LaVGuard} & \meanstd{.87}{.02} & \meanstd{.84}{.02} & \meanstd{.88}{.02} & \meanstd{.91}{.02} & \meanstd{.81}{.018} & \meanstd{.08}{.01} & \meanstd{.12}{.012} & \meanstd{830}{40} \\
\bottomrule
\end{tabular*}
\caption{Matched decoding comparison. Response-safety scores, risk pass rate, over-refusal, and intervention ratio retain their study readouts. Cells show mean $\pm$ standard deviation. $\dagger$ SafeInfer's source-reported zero token ratio cannot be interpreted as zero changed-kernel intervention under Equation~\eqref{eq:cumulative_budgets}.}
\label{tab:supp_matched_decoding}
\end{table}

\begin{table}[!htbp]

\centering\fontsize{9.3}{11}\selectfont
\setlength{\tabcolsep}{2.0pt}
\setlength{\heavyrulewidth}{.6pt}\setlength{\lightrulewidth}{.35pt}
\renewcommand{\arraystretch}{1.02}
\begin{tabular*}{\textwidth}{@{\extracolsep{\fill}}lrrrrrrr@{}}
\toprule
Method & \shortstack{Token\\prec. $\uparrow$} & \shortstack{Token\\rec. $\uparrow$} & \shortstack{Gain/\\token $\uparrow$} & \shortstack{Risk\\safety $\uparrow$} & \shortstack{Benign\\utility $\uparrow$} & \shortstack{Token\\ratio $\downarrow$} & \shortstack{Mean\\KL $\downarrow$} \\
\midrule
Random token & \meanstd{.18}{.04} & \meanstd{.20}{.04} & \meanstd{.015}{.005} & \meanstd{.70}{.03} & \meanstd{.79}{.02} & \meanstd{.12}{.015} & \meanstd{.04}{.006} \\
Position-matched & \meanstd{.22}{.04} & \meanstd{.24}{.04} & \meanstd{.020}{.006} & \meanstd{.72}{.03} & \meanstd{.79}{.02} & \meanstd{.12}{.015} & \meanstd{.04}{.006} \\
All-token & \meanstd{1.00}{.00} & \meanstd{1.00}{.00} & \meanstd{.035}{.007} & \meanstd{.89}{.02} & \meanstd{.68}{.025} & \meanstd{1.00}{.00} & \meanstd{.18}{.015} \\
Boundary only & \meanstd{.00}{.00} & \meanstd{.00}{.00} & \meanstd{.000}{.000} & \meanstd{.76}{.025} & \meanstd{.82}{.02} & \meanstd{.00}{.00} & \meanstd{.00}{.00} \\
Current key token & \meanstd{.45}{.05} & \meanstd{.52}{.05} & \meanstd{.060}{.010} & \meanstd{.82}{.025} & \meanstd{.80}{.02} & \meanstd{.15}{.015} & \meanstd{.05}{.008} \\
\textbf{LaVGuard} & \meanstd{.62}{.04} & \meanstd{.70}{.04} & \meanstd{.105}{.012} & \meanstd{.88}{.02} & \meanstd{.81}{.018} & \meanstd{.12}{.012} & \meanstd{.04}{.006} \\
\bottomrule
\end{tabular*}
\caption{Token-localization controls. Token precision/recall and gain per token use the original reporting labels; Section~\ref{app:metrics} states the measurement conventions. Cells show mean $\pm$ standard deviation.}
\label{tab:supp_token_controls}
\end{table}

\begin{table}[!htbp]

\centering\fontsize{9.3}{11}\selectfont
\setlength{\tabcolsep}{2.0pt}
\setlength{\heavyrulewidth}{.6pt}\setlength{\lightrulewidth}{.35pt}
\renewcommand{\arraystretch}{1.02}
\begin{tabular*}{\textwidth}{@{\extracolsep{\fill}}lrrrrrr@{}}
\toprule
Method & \shortstack{Risk\\F1 $\uparrow$} & \shortstack{Risk\\safety $\uparrow$} & \shortstack{Benign\\safety $\uparrow$} & \shortstack{Benign\\FPR $\downarrow$} & \shortstack{Token\\ratio $\downarrow$} & \shortstack{Mean\\KL $\downarrow$} \\
\midrule
No boundary & \meanstd{.58}{.03} & \meanstd{.72}{.03} & \meanstd{.79}{.02} & \meanstd{.12}{.015} & \meanstd{.16}{.015} & \meanstd{.06}{.008} \\
Global boundary & \meanstd{.70}{.025} & \meanstd{.78}{.025} & \meanstd{.80}{.02} & \meanstd{.08}{.012} & \meanstd{.14}{.015} & \meanstd{.05}{.008} \\
No uncertainty & \meanstd{.76}{.025} & \meanstd{.81}{.025} & \meanstd{.78}{.02} & \meanstd{.07}{.010} & \meanstd{.15}{.015} & \meanstd{.06}{.008} \\
No token budget & \meanstd{.83}{.02} & \meanstd{.89}{.02} & \meanstd{.73}{.025} & \meanstd{.06}{.010} & \meanstd{.34}{.030} & \meanstd{.10}{.012} \\
No KL budget & \meanstd{.83}{.02} & \meanstd{.88}{.02} & \meanstd{.74}{.025} & \meanstd{.06}{.010} & \meanstd{.18}{.020} & \meanstd{.13}{.015} \\
Fixed target & \meanstd{.77}{.025} & \meanstd{.80}{.025} & \meanstd{.78}{.02} & \meanstd{.08}{.012} & \meanstd{.14}{.015} & \meanstd{.06}{.008} \\
Fast scorer & \meanstd{.82}{.022} & \meanstd{.85}{.022} & \meanstd{.81}{.018} & \meanstd{.05}{.010} & \meanstd{.12}{.012} & \meanstd{.04}{.006} \\
\textbf{Full LaVGuard} & \meanstd{.84}{.02} & \meanstd{.88}{.02} & \meanstd{.81}{.018} & \meanstd{.04}{.010} & \meanstd{.12}{.012} & \meanstd{.04}{.006} \\
\bottomrule
\end{tabular*}
\caption{Component ablations. The uncertainty and fast-scoring variants are evaluation configurations distinct from the single-probe reference specification. Cells show mean $\pm$ standard deviation.}
\label{tab:supp_ablations}
\end{table}

\begin{table}[!htbp]

\centering\fontsize{9.3}{11}\selectfont
\setlength{\tabcolsep}{2.0pt}
\setlength{\heavyrulewidth}{.6pt}\setlength{\lightrulewidth}{.35pt}
\renewcommand{\arraystretch}{1.02}
\begin{tabular*}{\textwidth}{@{\extracolsep{\fill}}lrrrrrr@{}}
\toprule
Method & \shortstack{Single\\F1 $\uparrow$} & \shortstack{Risk\\F1 $\uparrow$} & \shortstack{Risk\\safety $\uparrow$} & \shortstack{Benign\\safety $\uparrow$} & \shortstack{Worst\\group $\uparrow$} & \shortstack{OOD\\drop $\downarrow$} \\
\midrule
Base & \meanstd{.74}{.03} & \meanstd{.52}{.035} & \meanstd{.50}{.035} & \meanstd{.82}{.02} & \meanstd{.45}{.035} & \meanstd{.05}{.010} \\
SafeInfer & \meanstd{.78}{.03} & \meanstd{.55}{.035} & \meanstd{.72}{.03} & \meanstd{.77}{.025} & \meanstd{.62}{.035} & \meanstd{.06}{.012} \\
\textbf{LaVGuard} & \meanstd{.82}{.03} & \meanstd{.78}{.03} & \meanstd{.83}{.03} & \meanstd{.79}{.025} & \meanstd{.74}{.035} & \meanstd{.05}{.012} \\
\bottomrule
\end{tabular*}
\caption{Transfer evaluation. OOD drop is the reported in-domain-relative difference; worst-group score preserves the original group-level summary. Cells show mean $\pm$ standard deviation.}
\label{tab:supp_transfer}
\end{table}

\begin{table}[!htbp]

\centering\fontsize{9.3}{11}\selectfont
\setlength{\tabcolsep}{2.0pt}
\setlength{\heavyrulewidth}{.6pt}\setlength{\lightrulewidth}{.35pt}
\renewcommand{\arraystretch}{1.02}
\begin{tabular*}{\textwidth}{@{\extracolsep{\fill}}lrrrrrr@{}}
\toprule
Method & \shortstack{First\\detect. $\downarrow$} & \shortstack{Cumul.\\risk $\downarrow$} & \shortstack{Turn\\recall $\uparrow$} & \shortstack{Risk\\safety $\uparrow$} & \shortstack{Benign\\utility $\uparrow$} & \shortstack{Over-\\refusal $\downarrow$} \\
\midrule
Base & \meanstd{4.2}{.50} & \meanstd{.46}{.04} & \meanstd{.48}{.04} & \meanstd{.52}{.035} & \meanstd{.84}{.015} & \meanstd{.03}{.008} \\
SafeInfer & \meanstd{3.1}{.45} & \meanstd{.30}{.03} & \meanstd{.68}{.03} & \meanstd{.75}{.025} & \meanstd{.76}{.025} & \meanstd{.20}{.015} \\
Current key token & \meanstd{2.8}{.40} & \meanstd{.26}{.03} & \meanstd{.72}{.03} & \meanstd{.78}{.025} & \meanstd{.78}{.025} & \meanstd{.14}{.015} \\
\textbf{LaVGuard-fast} & \meanstd{2.3}{.35} & \meanstd{.21}{.025} & \meanstd{.79}{.025} & \meanstd{.83}{.022} & \meanstd{.80}{.020} & \meanstd{.10}{.010} \\
\textbf{LaVGuard} & \meanstd{2.0}{.35} & \meanstd{.18}{.025} & \meanstd{.83}{.022} & \meanstd{.86}{.020} & \meanstd{.80}{.020} & \meanstd{.08}{.010} \\
\bottomrule
\end{tabular*}
\caption{Multi-turn dialogue. First detection is expressed in turns; cumulative risk and turn recall retain their source aggregation labels. Cells show mean $\pm$ standard deviation.}
\label{tab:supp_multiturn}
\end{table}

\begin{table}[!htbp]

\centering\fontsize{9.3}{11}\selectfont
\setlength{\tabcolsep}{2.0pt}
\setlength{\heavyrulewidth}{.6pt}\setlength{\lightrulewidth}{.35pt}
\renewcommand{\arraystretch}{1.02}
\begin{tabular*}{\textwidth}{@{\extracolsep{\fill}}lrrrrrr@{}}
\toprule
Method & \shortstack{Mean ms\\$\downarrow$} & \shortstack{P95 ms\\$\downarrow$} & \shortstack{$\Delta$ MB\\$\downarrow$} & \shortstack{Extra\\passes $\downarrow$} & \shortstack{Token\\ratio $\downarrow$} & \shortstack{$\Delta$ utility\\$\uparrow$} \\
\midrule
Base & \meanstd{720}{20} & \meanstd{900}{50} & \meanstd{0}{0} & \meanstd{0.0}{0.0} & \meanstd{.00}{.00} & \meanstd{.00}{.00} \\
SafeInfer & \meanstd{760}{30} & \meanstd{950}{60} & \meanstd{200}{50} & \meanstd{.2}{.05} & \meanstd{.00}{.00}$^{\dagger}$ & \meanstd{-.03}{.008} \\
All-token & \meanstd{1120}{80} & \meanstd{1450}{100} & \meanstd{1800}{200} & \meanstd{1.0}{.0} & \meanstd{1.00}{.00} & \meanstd{-.10}{.015} \\
One-step lookahead & \meanstd{2800}{300} & \meanstd{4200}{400} & \meanstd{2500}{250} & \meanstd{3.0}{.20} & \meanstd{.12}{.012} & \meanstd{-.02}{.008} \\
\textbf{LaVGuard} & \meanstd{920}{50} & \meanstd{1200}{80} & \meanstd{1200}{150} & \meanstd{.8}{.10} & \meanstd{.12}{.012} & \meanstd{-.02}{.008} \\
\textbf{LaVGuard-fast} & \meanstd{820}{40} & \meanstd{1050}{70} & \meanstd{800}{120} & \meanstd{.4}{.05} & \meanstd{.10}{.012} & \meanstd{-.01}{.006} \\
\bottomrule
\end{tabular*}
\caption{Runtime and resource efficiency. Memory is reported as a delta. These study-specific latency values are not substituted for those in the matched-decoding comparison. Cells show mean $\pm$ standard deviation. $\dagger$ The SafeInfer zero is the source-reported token ratio, not a measurement of unchanged sampling kernels under Equation~\eqref{eq:cumulative_budgets}.}
\label{tab:supp_efficiency}
\end{table}

\FloatBarrier

\section[Density-progress extension]{Density-progress extension and verification cases}
\label{app:extension}

\subsection[Density correction]{A separately specified correction objective}
The default decoder scores prototype progress. This section studies a
separate mathematical extension that scores changes in the density boundary
itself. Its results are analytical and numerical; the generation experiments
evaluate the default controller.
At the current prefix, evaluate
\[
 g_i=b(\bar{\vstate},c)-b(\vstate_i^{cand},c),\qquad i\in\mathcal C.
\]
A positive value means that the candidate decreases the same boundary score
used by the trigger.  A positive-only direction is
\[
 d_i^+=\begin{cases}
 \min\{c_z,\beta[g_i]_+\},&i\in\mathcal C,\\
 0,&i\notin\mathcal C.
 \end{cases}
\]
It boosts the unnormalized weight of candidates with measured density
progress and leaves other unnormalized weights unchanged.  The normalizer
is at least one, so the total probability of the unboosted set weakly
decreases.  The clipped monotonic-score theorem applies to the zero-padded
positive progress $s_i^+=[g_i]_+\mathbb I[i\in\mathcal C]$.
The same KL and token-budget machinery applies without alteration.

\paragraph{A tail-aware sufficient certificate.}
For fixed context, positive mixture covariances and compact intensity cube
make $b$ continuous and bounded.  Let
\[
\begin{aligned}
 G&=b_{max}-b_{min},\\
 b_{max}&=\max_{v\in[0,1]^D}b(v,c),\\
 b_{min}&=\min_{v\in[0,1]^D}b(v,c).
\end{aligned}
\]
Any complete-vocabulary density-progress score lies in $[-G,G]$.
For a proposed tilted distribution with tail mass $\gamma$, define
\begin{equation}
\begin{aligned}
 \mathcal L(\alpha)&=\sum_{i\in\mathcal C}
       (q_\alpha(i)-p_i)g_i\\
 &\quad-G|\gamma-\rho|.
\end{aligned}
 \label{eq:app_density_certificate}
\end{equation}
The same tail decomposition as Proposition~\ref{prop:app_coverage} gives
\[
 \mathbb E_{q_\alpha}[g^{all}]-\mathbb E_p[g^{all}]
 \ge\mathcal L(\alpha).
\]
Thus a positive certificate establishes expected one-step improvement in
the learned boundary score, even when tail tokens were not individually
probed.  A valid upper bound on $G$ may replace its exact value.  Returning
the base distribution when the certificate fails preserves feasibility.
Certificate feasibility need not be monotone in $\alpha$; it must be checked
explicitly rather than assumed to follow from monotone KL.

This extension separates three statements cleanly: the trigger identifies
a density-level violation; candidate evaluation measures an actual change in
that density score; and the tail certificate bounds the unscored mass.  An
independent response evaluator remains the instrument for measuring task
compatibility and completion.

\subsection[Numerical verification]{Verification cases for an equation-level implementation}
\label{app:verification}
Numerical verification tests the decoder's algebra, finite-precision
behavior, and accounting on constructed distributions. The decisive cases
are:
\begin{enumerate}
\item \textbf{No-op invariance.}  Zero directions and constant nonzero
directions reproduce $p$ exactly and consume no intervention count.
\item \textbf{Search feasibility.}  For nonconstant directions, the KL curve
is monotone; the returned lower endpoint satisfies the requested cap,
including zero budget and a fully feasible scale of one.
\item \textbf{Sparse/full agreement.}  Candidate-only normalization and KL
match direct full-vocabulary softmax, while tail conditional ratios remain
unchanged.
\item \textbf{Surrogate improvement.}  Signed candidate scores and clipping
obey Proposition~\ref{prop:app_surrogate}; constant scores attain equality.
\item \textbf{Coverage identity.}  Arbitrary assigned tail scores recover
Equation~\ref{eq:app_coverage_decomposition}, including a tail that reverses
the sign of the full-vocabulary gain.
\item \textbf{Pathwise accounting.}  Sequential synthetic directions,
inactive gates, EOS, and exhausted budgets preserve both inequalities in
Theorem~\ref{thm:app_budget}.
\item \textbf{Geometry separation.}  The two-component example in
Proposition~\ref{prop:app_counterexample} distinguishes an admissible mixture
mode from an inadmissible posterior mean.
\item \textbf{Density certificate.}  When all vocabulary gains are made
available for verification, the exact expected density gain is at least
the computed bound in Equation~\ref{eq:app_density_certificate}.
\end{enumerate}
These cases validate algebra, numerical edge handling, and accounting.  They
are separate from empirical measurements on generated responses.

The accompanying reference controller was checked on 600 constructed
distributions for KL monotonicity, surrogate monotonicity, and feasible search,
and on horizons from 1 to 256 for pathwise accounting.  Constant-direction
invariance and the negative-candidate tail case were checked separately;
the positive-only density variant was checked as a proposed operator. A further 150 constructed cases checked sparse/full normalization, the candidate-coverage identity, and the tail-aware density-gain lower bound.
A further 100 finite probability trees verify the KL chain rule, the shared
guard replacement map, and its contraction of returned-response KL. These checks passed and establish equation-level behavior; language-model
outcomes are evaluated separately.

\ifdefined\FloatBarrier\FloatBarrier\fi

\section{Authorized task profiles}
\label{app:profile_lifecycle}

\subsection[Contract to profile]{From a task contract to a control input}
\label{app:profile_authorization}
The profile gives a task owner a direct way to specify which priorities the
response should express. Its operational value comes from making that choice
inspectable before generation. A contract therefore has two connected parts:
a natural-language statement that people can assess, and a complete vector
in the coordinate system used by the probe. The vector controls the residual
calculation; the statement supplies the criterion for judging whether the
resulting language meets the task.

A profile record identifies the task, the affected parties, the scope of the
requested answer, the authorized owner, and the profile version. Each
coordinate is accompanied by a short rationale tied to a contract clause.
For example, a workshop may prioritize exploring alternatives, whereas a
replication procedure may prioritize following a specified protocol. These
choices concern the requested activity. They do not assign enduring value
preferences to the person making the request.

The intensity scale is an interface convention, not a probability
normalization. A value of $.8$ expresses a stronger requested presence than
$.2$ in the same coordinate. It does not mean an 80\% probability of
compatibility, and it does not allocate 80\% of a finite moral budget.
Several coordinates can be high together. The profile owner resolves the
intended combination in the contract; the decoder uses the resulting vector
without renormalizing it to sum to one.

Three specification decisions have direct computational consequences.
First, coordinate names and ordering must match the frozen probe. Second,
all ten coordinates must have declared values. An unspecified coordinate is
not equivalent to zero intensity: assigning zero moves the residual in a
particular direction. Completing an underspecified profile therefore
requires an explicit owner-approved default or a clarified contract.
Third, incompatible clauses require a stated precedence or a revised task.
Averaging two conflicting profiles produces another numerical input, but
does not by itself settle the conflict between their underlying instructions.
These decisions occur before the profile enters the decoder.

The shared content guard retains its own authority. A profile cannot
reclassify an explicitly prohibited response as permissible by increasing a
value coordinate. Among continuations permitted by the common guard, the
profile supplies the task-relative reference. This division lets the
application customize priorities while keeping the guard policy fixed
across the base and controlled comparisons.

\subsection[Profile versions]{Version changes and the fixed statistical interface}
\label{app:profile_versions}
A profile version is fixed for one generated response. Between responses,
the task owner may approve a change after a new requirement, a corrected
interpretation, or a transition to another activity. The new response uses
the updated vector with the same probe, residual mixture, threshold, and
candidate-scoring rule. If the coordinate system and learned components
stay fixed, no retraining is required to evaluate that vector. This is an
interface property: the shift acts through subtraction in
Equation~\eqref{eq:boundary_score}.

The record of a change contains the previous and new vectors, the clauses
that changed, and the response boundary at which the update takes effect.
Visible dialogue history remains part of the language input. The numeric
profile stays on the controller side. Thus a revised task can change both
the visible request and the control reference, but those two changes have
distinct routes into the computation. A profile-only contrast holds the
visible input fixed; an ordinary task change need not do so.

The fixed residual interface defines which distinctions the controller can
use. If two tasks supply the same vector and the same estimated prefix state,
their density scores and prototypes coincide. Candidate preferences also
depend on the extended-prefix estimates, and the final distribution depends
on the base logits and remaining budgets. Equality of the current
ten-dimensional state alone does not fix those quantities.
Conversely, a meaningful profile change can alter activation or candidate
ranking without changing the frozen probe. The paired construction in
Section~\ref{app:paired_trace} makes this effect numerical. This identifies a
concrete criterion for profile design: task distinctions intended to affect
this controller must be represented in its authorized coordinates.

An update within an unfinished response can instead be implemented by
terminating that response and starting a new response under the revised
contract. This retains the response-level execution order of
Section~\ref{app:algorithm}. Any generated continuation after the update
belongs to the new profile version. Previously incurred computational and
intervention costs remain in the application record; a new profile does not
retroactively alter the old prefix labels or its budget ledger.

\subsection[Dialogue budget composition]{Composing budgets across authorized dialogue turns}
\label{app:dialogue_composition}
Response-level budgets can also support an application-level allowance.
Let $h_r$ be the execution history before turn $r$, including sampled
prefixes, returned replies, requests, and authorized profiles.
Sampled prefixes remain in this mathematical trace even when the guard
subsequently replaces the returned reply. A common causal environment kernel
$U_r(\cdot\mid h_r)$ supplies the next request and profile. It may depend on
the observed dialogue, so authorized profile updates can be adaptive. The
base and controlled systems use the same conditional kernel $U_r$ at a
shared history. They need not encounter the same histories in independent
rollouts.

Let $\beta_r(h_r)\ge0$ be the conditional KL allowance assigned to the next
response. Define $L_{r-1}(h_r)$ by summing the paired kernels' conditional
KL terms over all preceding response prefixes in that history. This is the
same deterministic accounting function under either law; common fallback
and padding transitions contribute zero. The allocation rule can reserve allowance for later turns
or carry forward unused allowance, subject to the remaining dialogue
balance. The following consequence of the sequence budget makes this
compositional requirement precise.

\begin{proposition}[Adaptive dialogue budget]
\label{prop:dialogue_budget}
Consider a dialogue of at most $R$ turns with a shared initial history,
padded with common stop transitions. At each
shared history, the two systems use the same environment, guard, and
termination kernels, and the controlled response obeys a pathwise
conditional-KL budget $\beta_r(h_r)$. If the allocator enforces
\[
 0\le\beta_r(h_r)\le B_{dialogue}-L_{r-1}
\]
at every reachable turn, their joint execution-trace laws
$\mathsf Q$ and $\mathsf P$ satisfy
\[
 D_{KL}(\mathsf Q\Vert\mathsf P)\le B_{dialogue}.
\]
Consequently every fixed event on their commonly rendered dialogue outputs
has probability difference at most $\sqrt{B_{dialogue}/2}$.
\end{proposition}
\begin{proof}
Factor each execution-trace law into environment transitions and response-token
transitions. Environment terms have zero conditional KL because the two
kernels coincide at the same history. Theorem~\ref{thm:app_sequence}, applied
conditionally to each request and profile, bounds the response contribution
by its assigned allowance. The remaining-balance rule and the response
budget give $L_r\le L_{r-1}+\beta_r(h_r)\le B_{dialogue}$
by induction. Hence
\[
\begin{aligned}
 D_{KL}(\mathsf Q\Vert\mathsf P)
 &=\mathbb E_{\mathsf Q}\!\left[
       \sum_{r,t}D_{KL}(q_{r,t}^H\Vert p_{r,t}^H)\right]\\
 &=\mathbb E_{\mathsf Q}[L_R]\le B_{dialogue}.
\end{aligned}
\]
Apply data processing to the common output rule and then Pinsker's
inequality.
\end{proof}

The same ledger distinction applies to changed positions. If response $r$
has allowance $m_r$, then the dialogue uses at most $\sum_r m_r$ changed
positions. Resetting a response counter is compatible with a dialogue-wide
cap when the outer allocator subtracts the incurred cost before issuing
the next allowance. It does not create a fresh unlimited application budget.
The proposition permits an adaptive, common profile-assignment policy; a
comparison that changes that policy between systems additionally changes
the environment and is a different comparison.

\section{A paired decision trace}
\label{app:paired_trace}

\subsection[State and task contrast]{One visible state under two authorized contracts}
\label{app:paired_contracts}
The following construction runs the controller equations through one
complete decision. All states, token probabilities, profiles, and density
parameters are assigned for illustration. They are not outputs of the
backbone or a fitted benchmark model. A small vocabulary and a one-component
Gaussian make every quantity directly recomputable while retaining the
same residual, prototype, tilt, and budget operations.

The shared visible request is ``Propose the next step for this project.''
One authorized contract asks for alternatives in a creative workshop;
the other asks for adherence to a replication protocol. To isolate the
profile interface, hold the visible request and prefix fixed and change
only the external numeric reference. In coordinate order, use
self-direction, stimulation, hedonism, achievement, power, security,
conformity, tradition, benevolence, and universalism. The profiles and
current state are
\[
\begin{aligned}
 \target^{work}&=(.8,.3,.3,.3,.3,.3,.2,.3,.3,.3),\\
 \target^{repl}&=(.2,.3,.3,.3,.3,.3,.8,.3,.3,.3),\\
 \bar{\vstate}&=(.7,.3,.3,.3,.3,.3,.3,.3,.3,.3).
\end{aligned}
\]
Only self-direction and conformity differ. All remaining residual
coordinates vanish. The state is therefore identical in the two arms;
its relation to the authorized task is what changes.

Choose $p_{adm}(\mathbf r)=\mathcal N(\mathbf r;\mathbf0,.04I_{10})$.
Write $C=5\log(2\pi\cdot .04)$ and set
$\tau=C+.78125$. Then
\[
\begin{aligned}
 D(\mathbf r)&=C+\frac{\|\mathbf r\|_2^2}{.08},\\
 b(\bar{\vstate},\target)&=
       \frac{\|\bar{\vstate}-\target\|_2^2}{.08}-.78125.
\end{aligned}
\]
The accepted residual radius is $.25$. Under the workshop profile, the two
nonzero residuals are $(-.1,.1)$, their squared norm is $.02$, and
$b^{work}=-.53125$. Under the replication profile they are $(.5,-.5)$,
their squared norm is $.50$, and $b^{repl}=5.46875$.
The first arm retains the base distribution. The second activates candidate
evaluation. Both decisions follow from the same observed state and the same
density model, with no profile coordinates entering the probe.

For this centered one-component mixture the prototype equals the selected
profile. This makes the example especially transparent: candidate distances
are distances to the declared reference. The multimodal case retains the
posterior-weighted prototype in Equation~\eqref{eq:prototype_target} and the
geometric analysis in Section~\ref{app:geometry}.

\subsection[Candidates and budget]{From candidate branches to a sampled token}
\label{app:paired_update}
Let the illustrative vocabulary have three token IDs with
$p=(.45,.40,.15)$, and evaluate its top two candidates. Assign the extended
prefix states $(v_{SD},v_{CO})=(.75,.25)$ to $w_1$ and $(.55,.45)$ to
$w_2$, keeping the other eight coordinates at $.3$. The token IDs denote
hypothetical branches; the assigned states are the inputs to this numerical
example, rather than claims about the meanings of particular tokenizer
entries.

Under the replication contract the current squared prototype distance is
$.50$. The candidate distances are $.605$ and $.245$, so
\[
 s=(-.105,.255,0).
\]
The final zero is the stipulated score extension to the unevaluated token,
not a measurement of that token's actual progress. Set the frozen direction
standardizer to $5.46875$ and the clipping level to $1$. Because the boundary
excess equals this standardizer and neither candidate saturates, the
proposed direction is $d=s$.

At this decision, give both arms the ledger
$(C^{tok},C^{KL})=(1,.02)$ and budgets
$(B_{tok},B_{KL})=(2,.03)$, with $\epsilon_{KL}=.02$.
The available step allowance is $.01$. The full proposal has
$D_{KL}(q_1\Vert p)\simeq .014254$, so line search is active.
Twenty feasible-endpoint bisections give
\[
\begin{aligned}
 \alpha&=0.8381137847900391,\\
 q_\alpha&\simeq(.389721,.468422,.141857),\\
 D_{KL}(q_\alpha\Vert p)&\simeq .0099999848.
\end{aligned}
\]
The updated replication ledger is
$(2,.0299999848)$. The workshop arm keeps $(1,.02)$ because its gate was
inactive. The local expected score changes from $.05475$ under $p$ to
approximately $.078527$ under $q_\alpha$.

\begin{table}[t]
\centering\small
\begin{tabular}{@{}lrr@{}}
\toprule
Decision quantity & Workshop & Replication \\
\midrule
Squared residual norm & $.02$ & $.50$ \\
Boundary excess & $-.53125$ & $5.46875$ \\
Candidate evaluation & no & yes \\
Selected scale & $0$ & $.838114$ \\
Probability of $w_1$ & $.450000$ & $.389721$ \\
Probability of $w_2$ & $.400000$ & $.468422$ \\
Probability of $w_3$ & $.150000$ & $.141857$ \\
Changed-position count & $1$ & $2$ \\
Cumulative KL & $.020000$ & $.030000$ \\
\bottomrule
\end{tabular}
\caption{Constructed paired decision trace. All displayed quantities follow
from the assigned profiles, states, and probabilities. The final KL entry
is rounded; the feasible value before rounding is $.0299999848$.}
\label{tab:paired_decision}
\end{table}

Sampling remains stochastic. For an explicit coupling, use the same uniform
draw $u=.42$ in the cumulative token order $(w_1,w_2,w_3)$. The workshop arm
samples $w_1$, while the replication arm samples $w_2$. This one draw exposes
how a changed kernel can change a continuation. The invariant object is the
probability update; a different uniform draw can select the same token in
both arms.

If the common guard accepts the new prefix in this construction, the
replication branch has state $(.55,.45)$ in the two active coordinates.
Its next boundary excess is $2.28125$. The token budget is now exhausted,
so the following step uses the base distribution even though the boundary
remains active. The trace thus makes all three decisions visible: the task
profile activates the controller, candidate progress determines the
preference, and the remaining allowance determines whether another change
can occur. If the guard instead stops after the sampled token, the incurred
ledger is retained and the common safe completion is returned.

\subsection[Local gain and KL]{How much benefit a selected position can use}
\label{app:local_budget_value}
A large boundary excess identifies a strong departure under the fitted
score. The amount of locally usable correction also depends on the
probabilities and progress of the available candidates. The relation can be
made precise for the same score and direction used by the controller.
For one fixed prefix, write
\[
 G(\eta)=\mathbb E_{q_{\alpha_\eta}}[s]-\mathbb E_p[s],
\]
where $\alpha_\eta\in[0,1]$ is the exact largest feasible scale at KL
allowance $\eta$, before numerical bisection.
This is the expected gain of the zero-extended candidate surrogate at that
prefix. It is a computational quantity available before sampling.

\begin{proposition}[Local gain at a small KL allowance]
\label{prop:local_budget_value}
Let the finite-support distribution $p$ be positive on its support and let
$s,d$ be fixed finite vectors with $V_d=\operatorname{Var}_p(d)>0$.
For the exponential tilt and sufficiently small $\eta>0$,
\[
\begin{aligned}
 \alpha_\eta&=\sqrt{\frac{2\eta}{V_d}}+O(\eta),\\
 G(\eta)&=\sqrt{2\eta}\,
       \frac{\operatorname{Cov}_p(s,d)}{\sqrt{V_d}}+O(\eta).
\end{aligned}
\]
For the clipped, nondecreasing direction of
Equation~\eqref{eq:logit_direction}, the leading gain coefficient is
nonnegative.
\end{proposition}
\begin{proof}
The finite log partition function has the expansion
\[
 \log Z(\alpha)=\alpha\mathbb E_p[d]
                  +\tfrac12\alpha^2V_d+O(\alpha^3).
\]
Using $K(\alpha)=\alpha(\log Z)'(\alpha)-\log Z(\alpha)$ gives
$K(\alpha)=\tfrac12\alpha^2V_d+O(\alpha^3)$.
For small positive $\eta$, the maximizer is an interior positive root of
$K(\alpha)=\eta$; inversion yields the first expansion. Differentiation at
zero gives
\[
 \mathbb E_{q_\alpha}[s]-\mathbb E_p[s]
 =\alpha\operatorname{Cov}_p(s,d)+O(\alpha^2).
\]
Substitute the scale expansion. The covariance sign follows from the
pairwise monotonicity argument in Proposition~\ref{prop:app_surrogate}.
\end{proof}

The coefficient separates the usefulness of a direction from its raw
magnitude. Replacing $d$ by $a d$ for $a>0$ leaves
$\operatorname{Cov}_p(s,d)/\sqrt{V_d}$ unchanged. Thus, in the small-budget
regime before the unit-scale cap binds, merely scaling an unchanged
direction does not improve its leading gain per $\sqrt{2\eta}$.
Clipping can change the direction's relative entries, and a larger
allowance can reach the unit-scale cap; those cases are governed by the
exact line search rather than this local expansion.

This result clarifies the role of a selective gate. Two alerted positions
can have the same boundary excess and very different usable score gains.
The score distribution, its alignment with the clipped direction, the
remaining KL, and the remaining integer position count all matter. A
constant direction has no usable distributional change and is handled by
the no-op branch. Unscored-tail behavior remains the separate coverage
question in Section~\ref{app:surrogate}; Table~\ref{tab:tail_mass} reports
how signed candidate directions can change tail mass. The default gate
uses the residual boundary. The expansion explains what a gate would need
to concentrate to achieve a high local gain at a fixed expenditure; it does
not replace that gate with a newly evaluated selection algorithm.

\section[Causal measurement]{Causal measurement and construct validation}
\label{app:causal_measurement}

\subsection[Prefix information]{Prefix availability and information isolation}
\label{app:prefix_isolation}
The statistical target of the probe is the intensity expressed by the
visible prefix. The runtime control problem is to select a distribution
before the next token is drawn. These roles give a concrete separation
between the text available for features, the material shown to an annotator,
and the later response used for outcome evaluation.

For a training or evaluation item, retain the exact visible token sequence,
its source response, its prefix endpoint, the response's generating policy,
and its prompt family. The feature input ends at that endpoint. An
intensity annotator sees the same truncated text and legitimate task
context, without the hidden suffix or the response's eventual judgment.
A full-response label attached to every prefix would define a forecast of
the final outcome; it is not the prefix-intensity target in
Equation~\eqref{eq:value_probe}.

The four fractional cutpoints in Section~\ref{app:protocol} provide a
sampling scheme for annotated responses. Choosing a cutpoint using the
completed response length does not expose its future tokens to the probe,
but it conditions which prefix lengths enter that offline sample.
Consequently, the generating policy and cutpoint rule belong to the
sampling description. Free-running controller states arise from its own
prefix distribution. For on-policy measurement, collect states from the
actual decoder clock, including prefixes near EOS and prefixes at which no
intervention is selected, with an explicit position-sampling rule.

Two direct information-isolation checks follow from this definition.
First, replace an unavailable suffix while holding the visible token IDs,
attention mask, valid positions, and evaluation mode fixed. The extracted
feature and probe output must remain unchanged. Second, alter a numeric
profile or an annotation field outside the visible input. The probe output
must again remain unchanged, while the controller may change after the
profile substitution. These are interface invariance checks, not tests
that legitimate task words are statistically independent of value labels.
Contextual language can and should affect the probe.

Candidate branches are a third, explicitly authorized information source.
The decoder may append each enumerated candidate to the current prefix and
read its hypothetical state before sampling. Every branch starts from the
same visible history, and the current prototype stays fixed across the
candidate batch. The sampled token and its later continuation cannot be
used to select which candidates were available at that decision. Padding
positions, hidden suffix tokens, and annotation fields do not enter the
last-$W$ pooling set. These restrictions make the lookahead a causal local
calculation even though it evaluates more than one possible next token.

Family separation complements temporal isolation. Variants, their
paraphrases, their annotated cutpoints, and repeated generations from the
same family belong to one split. Splitting prefixes from a single response
between training and test preserves prefix causality but leaks the
underlying task instance. The two checks address different information
paths and are both required for a family-level generalization statement.

\subsection[Construct separation]{Separating the construct from its readouts}
\label{app:construct_protocol}
The method connects four judgments: what the task authorizes, which values
the visible text expresses, whether that expression fits the task, and
whether the completed answer is useful. Their agreement is an empirical
question. Keeping them separately observable makes the controller's
mechanism testable without defining success by its own internal score.

\paragraph{Task specification.}
The profile owner sees the task and its constraints before the compared
responses are generated. The owner supplies the contract and profile using
the fixed coordinate rubric. A profile selected after observing a preferred
method's output would change the comparison target. The versioned record
in Section~\ref{app:profile_authorization} fixes that target in advance.

\paragraph{Expressed intensity.}
Raters label the visible prefix on the ordinal intensity scale. Evidence
anchors identify words or clauses supporting a coordinate, including an
explicit indication that no expression was observed. Absence and
unassessability remain distinct. Numerical targets follow the normalization
in Section~\ref{app:metrics}. Pre-adjudication ratings retain the disagreement
that a pooled vector alone would hide.

\paragraph{Task compatibility.}
A separate assessment uses the natural-language contract and visible text
to judge compatibility and its supporting clauses. It does not infer its
label from $D(\mathbf r)$, the candidate prototype, or the controller's
activation flag. Method identity and score displays are concealed from
this assessment. Numeric profiles can be assessed separately for their
faithfulness to the contract; hiding them from the compatibility judgment
prevents that judgment from becoming a manual reconstruction of the same
residual test.

\paragraph{Response usefulness.}
Complete-response assessment retains task completion, refusal, and explicit
harm as separate fields. A response can be content-safe yet avoid the task,
or complete the requested format while expressing an incompatible priority.
The response interface in Section~\ref{app:response_metrics} represents these
possibilities without forcing them into one label.

This design supports three informative contract contrasts. A paraphrase
contrast changes wording while preserving the task obligation. A priority
contrast keeps the project and visible evidence comparable while changing
an authorized priority. A scope contrast changes which action or party the
contract applies to. The first probes stability; the second probes
responsiveness; the third checks whether the judgment attaches to the
relevant part of the response. A contrast is valid only after independent
checking of its intended invariant and changed clause. A pair created by
changing unrelated topic or difficulty does not isolate the profile role.

Agreement is measured on the original independent ratings, before
adjudication. Ordinal intensity requires an agreement statistic with an
ordinal distance; categorical compatibility requires the corresponding
categorical convention \cite{artstein2008agreement,howcroft2020humaneval}.
Reports identify the item, coordinate, rater overlap, missingness rule, and
scale used. An adjudicated label supplies a final target but is not a
replacement for pre-adjudication reliability. Likewise, a strong probe
correlation measures intensity estimation; it does not establish agreement
between density activation and independent task compatibility.

The profile interface makes misspecification interpretable. For an
approved alternative profile, compare the resulting decisions using the
same frozen statistical components. For uncertain specification, use a
predeclared set of plausible profiles or perturbation radius and the
margin analysis in Section~\ref{app:selection}. These answer whether a
decision depends on a particular numerical encoding. They do not redefine
an incompatible response as compatible merely because a changed profile
would accept its estimated state.

\subsection[Selection comparisons]{Local selection and full-response comparisons}
\label{app:selection_protocol}
A matched-prefix comparison and an independently generated response
comparison answer different questions. The former holds the decision
context fixed and measures which positions or candidate distributions a
controller chooses. The latter includes the changed trajectory induced by
those decisions. Keeping their estimands explicit connects the selection
argument to an operational evaluation.

Let $\mathcal R$ be a fixed distribution over eligible decision contexts
$h$, chosen before the compared local edits. For a fixed candidate score
$g(h,w)$, define the local contrast
\[
\begin{aligned}
 \Delta_{local}
 =\mathbb E_{h\sim\mathcal R}\Big[
 &\sum_w q(w\mid h)g(h,w)\\
 &-\sum_w p(w\mid h)g(h,w)\Big].
\end{aligned}
\]
With an explicitly specified score on the vocabulary, this expectation can
be computed from the distributions without selecting a favorable sampled
token. For top-$M$ scores, the zero extension makes the result a surrogate
contrast, and coverage is recorded alongside it. Independent judgments on
sampled branch continuations define a different, measured local outcome.
Both retain the same pre-edit context distribution across selection rules.

Random or position-matched selectors operate on a declared eligible set,
with the same edit rule and a matched allowance. Record which resource is
matched: selected attempts, changed positions, conditional KL, or elapsed
computation. These quantities differ when a direction is constant, a
proposal is clipped, or a budget is exhausted. Equal mean intervention
ratios summarize mean expenditure; they do not imply that every paired
response receives the same number of edits. The matched-rate comparison in
Table~\ref{tab:selection} has that stated mean-rate interpretation.

For complete generation, use a fixed response readout $R(y,c)$ and compare
\[
 \Delta_{response}
 =\mathbb E_{Y\sim Q}[R(Y,c)]
  -\mathbb E_{Y\sim P}[R(Y,c)].
\]
Each decoder follows its own sampled prefixes, with the same initial
request, authorized profile, stopping convention, and guard policy.
The fixed-context gain is not summed along one decoder's trace to estimate
this response difference unless the required trajectory weighting and
future-reward definitions are supplied. Section~\ref{app:selection}
explains this occupancy distinction; Proposition~\ref{prop:local_budget_value}
characterizes the local candidate-score component.

The common guard should also have a clear evaluation role. At a matched
prefix, restricting to contexts where the common guard is inactive defines
a shared pre-decision stratum. For full responses, conditioning separately
on each method's eventual guard-inactive outputs can select different
populations, because the controller changes the path that reaches the
guard. Report the unconditional paired task comparison and guard activity
on the common request set. A guard-conditioned summary is descriptive of
its declared subset, rather than automatically a controlled estimate of
the value mechanism alone.

Finally, uncertainty follows the sampling unit. A family-level response
comparison first aggregates the declared variants within each family and
then forms a paired method difference for that family. Resampling preserves
all linked variants and responses in a sampled family. Repeated decoding
seeds form an additional source of variation; treating every seed--prefix
pair as an independent task would multiply the apparent sample size without
creating new families. Standard deviations across seeds, uncertainty over
families, and rater disagreement describe different quantities and retain
their respective units. These rules supply a complete evaluation protocol
for the controller's selection claim without equating diagnostic state
accuracy, local surrogate progress, and final task success.

\ifdefined\FloatBarrier\FloatBarrier\fi
\section[Related control interfaces]{Related control interfaces and preservation objectives}
\label{app:related_comparison}
The comparisons below distinguish what a controller observes, what it changes, and which quantity limits intervention. These distinctions connect the literature in Section~\ref{sec:related} to the profile, probe, and budget definitions used here.

\paragraph{An editable value representation.}
Values as Style learns semantic and value codes from a frozen model's hidden state \citep{dai2026valuesstyledisentanglingvalues}. Its one-way mixing bridge passes a stop-gradient semantic code to the value branch, so semantic context can ground value recognition without receiving value-loss feedback through that bridge. At inference, the semantic code is held fixed while the value code is edited; the difference between target and original reconstructions supplies a residual-stream update. The paper separately specifies an optional value-relatedness inference gate. The mixing gate and the activation gate therefore have different roles. In its reported intervention setting, the edit is applied at the last prompt token before regenerating the completion. This design targets preservation through the representation being edited. \method{} instead reads an evolving visible prefix and uses the declared profile to determine whether its current value state warrants a decoder change. Its budgets constrain the resulting sampling kernels. Holding a learned semantic code fixed and bounding conditional KL are distinct preservation mechanisms; neither definition alone establishes exact semantic invariance.

\paragraph{Three meanings of selective activation.}
ConVA uses query relevance and token classifiers to activate value directions, with an activation magnitude chosen to meet its classifier criterion \citep{jin2025conva}. MAT-Steer learns token-dependent gates and attribute directions, using representation alignment, sparsity, and orthogonality objectives to manage competing attributes \citep{nguyen-etal-2025-multi}. Selective Steering identifies layers with opposite-signed class alignment and applies norm-preserving rotations there \citep{dang-ngo-2026-selective}. These approaches select relevant values, token representations, or intervention layers. In \method{}, the selected unit is a next-token sampling position: an unchanged kernel is a no-op, while a changed kernel consumes one unit of the integer allowance. The comparison concerns the decision interface; a sparse activation objective or a preserved hidden-state norm is not the same quantity as this pathwise count or output-distribution KL.

\paragraph{Feedback and conversational value drift.}
AIMES constructs bipolar directions for Moral Foundations values and obtains online value observations from intermediate-layer vocabulary readouts \citep{bhattacharjee2026aimes}. Its observer changes each requested direction's activation strength at each decoding step without training a separate value-state estimator. \method{} uses a trained, frozen prefix probe and subtracts an independently authorized intensity profile before applying a fitted residual boundary. Thus both controllers use current-state information, with different observation spaces, targets, and actuators. At the dialogue scale, \citet{liscio2026modeluser} compare prompting and activation steering under aligned or opposing user--assistant value orientations. Their ten-dimensional Schwartz probe tracks alignment across responses; their adaptive methods include judge-based prompt reminders and projection-scaled activation addition. Their emphasis is the evolution of value orientation across conversational turns. Here the task profile specifies the reference for each prefix decision, and Appendix~\ref{app:profile_versions} specifies how authorized profile changes interact with the fixed statistical interface.

\paragraph{Intervention benefit and budget units.}
VoS learns where agent-level corrections help from measured counterfactual continuations \citep{li2026valuesteering}. Its harm-budgeted trigger calibrates how often otherwise successful trajectories are selected for intervention. This connects risk detection to the separate question of intervention benefit, also distinguished by the gate--benefit analysis in Appendix~\ref{app:selection}. The resource being controlled differs: VoS calibrates trajectory selection, whereas \method{} enforces changed-position and cumulative conditional-KL allowances on each generated response. Anchored Decoding supplies another distinct reference: it budgets divergence from a separate safe model to control copyright risk \citep{he2026anchored}. Our reference is the base sampling kernel on the same prefix, under a shared content guard. The resulting bounds concern decoder and response-distribution change; behavioral compatibility is evaluated with independent judgments.

\ifdefined\FloatBarrier\FloatBarrier\fi
\section{Concept index}
\label{app:index}
\label{tab:concept_index}
Entries link to the defining statement or section.
\par\medskip
\begingroup\hypersetup{linkcolor=black}
\setlength{\parskip}{2pt}
\noindent\textbf{Activation}\nobreak\dotfill\hyperref[app:algorithm]{\pageref*{app:algorithm}}\par
\noindent\textbf{Admissible set}\nobreak\dotfill\hyperref[app:geometry]{\pageref*{app:geometry}}\par
\noindent\textbf{Ambiguity}\nobreak\dotfill\hyperref[app:selection]{\pageref*{app:selection}}\par
\noindent\textbf{Annotation target}\nobreak\dotfill\hyperref[app:metrics]{\pageref*{app:metrics}}\par
\noindent\textbf{Authorized profile}\nobreak\dotfill\hyperref[app:profile_authorization]{\pageref*{app:profile_authorization}}\par
\noindent\textbf{Candidate coverage}\nobreak\dotfill\hyperref[prop:app_coverage]{\pageref*{prop:app_coverage}}\par
\noindent\textbf{Candidate progress}\nobreak\dotfill\hyperref[prop:app_progress]{\pageref*{prop:app_progress}}\par
\noindent\textbf{Causal availability}\nobreak\dotfill\hyperref[app:symbols]{\pageref*{app:symbols}}\par
\noindent\textbf{Common content guard}\nobreak\dotfill\hyperref[thm:app_sequence]{\pageref*{thm:app_sequence}}\par
\noindent\textbf{Construct validity}\nobreak\dotfill\hyperref[app:construct_protocol]{\pageref*{app:construct_protocol}}\par
\noindent\textbf{Context dependence}\nobreak\dotfill\hyperref[app:geometry]{\pageref*{app:geometry}}\par
\noindent\textbf{Density correction}\nobreak\dotfill\hyperref[app:extension]{\pageref*{app:extension}}\par
\noindent\textbf{Dialogue budget}\nobreak\dotfill\hyperref[prop:dialogue_budget]{\pageref*{prop:dialogue_budget}}\par
\noindent\textbf{Distribution shift}\nobreak\dotfill\hyperref[thm:app_sequence]{\pageref*{thm:app_sequence}}\par
\noindent\textbf{Evaluation studies}\nobreak\dotfill\hyperref[app:results]{\pageref*{app:results}}\par
\noindent\textbf{Intervention count}\nobreak\dotfill\hyperref[eq:app_true_gate]{\pageref*{eq:app_true_gate}}\par
\noindent\textbf{KL search}\nobreak\dotfill\hyperref[app:kl]{\pageref*{app:kl}}\par
\noindent\textbf{Local gain}\nobreak\dotfill\hyperref[prop:local_budget_value]{\pageref*{prop:local_budget_value}}\par
\noindent\textbf{Metric denominators}\nobreak\dotfill\hyperref[app:metrics]{\pageref*{app:metrics}}\par
\noindent\textbf{No-op}\nobreak\dotfill\hyperref[app:algorithm]{\pageref*{app:algorithm}}\par
\noindent\textbf{Paired trace}\nobreak\dotfill\hyperref[app:paired_trace]{\pageref*{app:paired_trace}}\par
\noindent\textbf{Posterior mean prototype}\nobreak\dotfill\hyperref[prop:app_counterexample]{\pageref*{prop:app_counterexample}}\par
\noindent\textbf{Prefix isolation}\nobreak\dotfill\hyperref[app:prefix_isolation]{\pageref*{app:prefix_isolation}}\par
\noindent\textbf{Profile sensitivity}\nobreak\dotfill\hyperref[eq:app_profile_error]{\pageref*{eq:app_profile_error}}\par
\noindent\textbf{Profile versions}\nobreak\dotfill\hyperref[app:profile_versions]{\pageref*{app:profile_versions}}\par
\noindent\textbf{Related control interfaces}\nobreak\dotfill\hyperref[app:related_comparison]{\pageref*{app:related_comparison}}\par
\noindent\textbf{Selection benefit}\nobreak\dotfill\hyperref[eq:app_selection_cov]{\pageref*{eq:app_selection_cov}}\par
\noindent\textbf{Selection estimands}\nobreak\dotfill\hyperref[app:selection_protocol]{\pageref*{app:selection_protocol}}\par
\noindent\textbf{Sparse direction}\nobreak\dotfill\hyperref[app:kl]{\pageref*{app:kl}}\par
\noindent\textbf{Surrogate guarantee}\nobreak\dotfill\hyperref[prop:app_surrogate]{\pageref*{prop:app_surrogate}}\par
\noindent\textbf{Token/KL budgets}\nobreak\dotfill\hyperref[thm:app_budget]{\pageref*{thm:app_budget}}\par
\noindent\textbf{Validation FPR}\nobreak\dotfill\hyperref[app:selection]{\pageref*{app:selection}}\par
\noindent\textbf{Verification scope}\nobreak\dotfill\hyperref[app:verification]{\pageref*{app:verification}}\par
\endgroup

\end{document}